\documentclass[preprint,review,authoryear,12pt]{elsarticle}
\usepackage[margin=2.5cm]{geometry}

\usepackage{amssymb}
\usepackage{amsmath}
\usepackage{booktabs}
\usepackage{listings}
\usepackage{pdflscape}
\usepackage{threeparttable}
\usepackage{url}

\begin{document}

\begin{frontmatter}

\title{Assessing the potential of PlanetScope satellite imagery to estimate particulate matter oxidative potential}

\author[1]{Ian Hough\corref{cor1}}
\author[2]{Loïc Argentier}
\author[3]{Ziyang Jiang}
\author[4]{Tongshu Zheng}
\author[3,5]{Mike Bergin}
\author[3,6]{David Carlson}
\author[1]{Jean-Luc Jaffrezo}
\author[7]{Jocelyn Chanussot\corref{cor2}}
\author[1]{Gaëlle Uzu\corref{cor2}}

\affiliation[1]{
    organization={Univ. Grenoble Alpes, CNRS, INRAE, IRD, Grenoble INP, IGE},
    city={Grenoble},
    postcode={38000},
    country={France}
}
\affiliation[2]{
    organization={Data Science Experts},
    city={Grenoble},
    postcode={38000},
    country={France}
}
\affiliation[3]{
    organization={Department of Civil and Environmental Engineering, Duke University},
    city={Durham},
    state={North Carolina},
    postcode={27708},
    country={USA}
}
\affiliation[4]{
    organization={Duke Kunshan University, Division of Natural and Applied Sciences},
    city={Kunshan},
    postcode={215316},
    country={China}
}
\affiliation[5]{
    organization={Duke Global Health Institute, Duke University},
    city={Durham},
    state={North Carolina},
    postcode={27708},
    country={USA}
}
\affiliation[6]{
    organization={Department of Biostatistics and Bioinformatics, Duke University},
    city={Durham},
    state={North Carolina},
    postcode={27708},
    country={USA}
}
\affiliation[7]{
    organization={Univ. Grenoble Alpes, Inria, CNRS, Grenoble INP, LJK},
    city={Grenoble},
    postcode={38000}, 
    country={France}
}

\cortext[cor1]{Corresponding author: ian.hough@univ-grenoble-alpes.fr}
\cortext[cor2]{Co-last authors}

\begin{abstract}
Oxidative potential (OP), which measures particulate matter's (PM) capacity to induce oxidative stress in the lungs, is increasingly recognized as an indicator of PM toxicity. Since OP is not routinely monitored, it can be challenging to estimate exposure and health impacts. Remote sensing data are commonly used to estimate PM mass concentration, but have never been used to estimate OP. In this study, we evaluate the potential of satellite images to estimate OP as measured by acellular ascorbic acid (OP\textsubscript{AA}) and dithiothreitol (OP\textsubscript{DTT}) assays of 24-hour PM\textsubscript{10} sampled periodically over five years at three locations around Grenoble, France. We use a deep convolutional neural network to extract features of daily 3~m/pixel PlanetScope satellite images and train a multilayer perceptron to estimate OP at a 1~km spatial resolution based on the image features and common meteorological variables. The model captures more than half of the variation in OP\textsubscript{AA} and almost half of the variation in OP\textsubscript{DTT} (test set R\textsuperscript{2} = 0.62 and 0.48, respectively), with relative mean absolute error (MAE) of about 32\%. Using only satellite images, the model still captures about half of the variation in OP\textsubscript{AA} and one third of the variation in OP\textsubscript{DTT} (test set R\textsuperscript{2} = 0.49 and 0.36, respectively) with relative MAE of about 37\%. If confirmed in other areas, our approach could represent a low-cost method for expanding the temporal or spatial coverage of OP estimates.
\end{abstract}

\begin{keyword}
air pollution \sep particulate matter \sep oxidative potential \sep oxidative burden \sep remote sensing \sep deep learning
\end{keyword}

\end{frontmatter}

\section{Introduction}

Particulate matter (PM) air pollution is a leading environmental health hazard that has been linked to adverse cardiovascular, respiratory, neurological, and perinatal effects \citep{WHO2021}. PM is a physically and chemically diverse mixture whose components vary widely over space and time and in toxicity. Current air quality guidelines aim to protect health by limiting the ambient mass concentration of two size fractions of PM, particles with aerodynamic diameter \textless 10~µm (PM\textsubscript{10}) and \textless~2.5~µm (PM\textsubscript{2.5}) \citep{WHO2021}. PM mass concentration and size are relatively easy to measure and are biologically relevant because they affect the exposure dose and how deeply inhaled particles penetrate into the lungs. But alternate metrics more closely related to PM toxicity might better predict health impacts and could help inform public health and emissions reduction policies.

Oxidative potential (OP) measures PM’s capacity to induce oxidative stress in the lungs. OP has been suggested as a possible summary metric for PM’s health impact because oxidative stress is one of the main biological pathways linking PM exposure to adverse health effects \citep{Feng2016, U.S.EPA2019, Al-Kindi2020, Leikauf2020}. OP is highly dependent on PM composition, and the chemical species and emissions sources that contribute most to OP are often not the main drivers of PM mass concentration \citep{Kelly2012, Daellenbach2020, Weber2021}. Some epidemiological studies found that OP was more strongly associated with cardiovascular and respiratory health than PM concentration, or that high OP magnified the relationship between PM concentration and health \citep{Bates2019, Gao2020, He2023}. However, other studies have found no clear association between OP and health, and the number of epidemiological studies examining OP exposure remains limited.

Efforts to evaluate OP's relevance as a health metric are limited by the lack of standardized protocols for measuring OP, which makes it difficult to compare results between studies \citep{Bates2019, He2023}, and the sparsity of measurements, which make it hard to estimate exposure \citep{Gao2020, He2023}. OP measurement is costly and is not routinely performed at air quality monitoring sites. This has led most epidemiological studies to estimate exposure based on OP measured over a limited time period at a central location, which risks underestimating or failing to detect health effects \citep{Zeger2000}.

Satellite observations are widely used to estimate ambient PM mass concentration at the earth's surface \citep{Chen2019, Di2019, VanDonkelaar2021, Wei2023a, Mandal2024}. A few recent studies have used satellite data to estimate PM composition, either directly \citep{Meng2018, Chau2020} or in combination with chemical transport modelling \citep{VanDonkelaar2019, Chen2020, Wei2023b}. But, to the best of our knowledge, no studies have explored using satellite data to estimate OP. Two types of OP indicators might be derivable from satellite data. First, changes in top-of-atmosphere reflectance over time may be related to temporal variation in PM composition. Second, surface features visible in satellite imagery (such as roads, vegetation, and building density) may be related to spatial variation in emissions sources or other factors that affect PM composition.

Most satellite-based models of PM mass concentration use aerosol optical depth (AOD), a measure of light extinction due to scattering and absorption by aerosols. Satellite-derived AOD may be a limited indicator of OP because it depends on both PM concentration and composition while OP is mainly related to PM composition. AOD is also limited in resolution, with current products having a maximum spatial resolution of about 1~km\textsuperscript{2} for a temporal resolution of 1~day. An alternative is to use lower-level satellite products such as top-of-atmosphere imagery. These contain more information than AOD and could avoid propagating errors introduced by the AOD retrieval process. Satellite imagery is also available at higher spatial resolutions than AOD, and so has the potential to capture finer spatial gradients over urban areas and other complex terrain. Several recent studies in China \citep{Shen2018, Liu2019, Yang2020, Fan2021, Wang2021, Yan2021, Yin2021, Yang2022, Tang2023} and Korea \citep{Choi2023} have used top-of-atmosphere reflectance to model PM concentration at up to 90~m spatial resolution, with some reporting slightly improved performance compared to models based on AOD.

PlanetScope satellite images provide global daily coverage with a 3~m/pixel resolution, and so could allow for very high-resolution air quality estimates. PlanetScope images were recently used to estimate daily PM2.5 mass concentration in Beijing and Delhi \citep{Zheng2020, Zheng2021, Jiang2022}. These studies used convolutional neural networks (CNNs), a deep learning architecture that performs very well on a variety of computer vision tasks \citep{Li2022, Yuan2020}. In light of these achievements, this study we explore the potential of deep learning and PlanetScope images to estimate daily PM\textsubscript{10} mass concentration and OP in Grenoble, France.

\section{Materials}

\subsection{Study area}

\begin{figure}[!t]
    \includegraphics[width=1\linewidth]{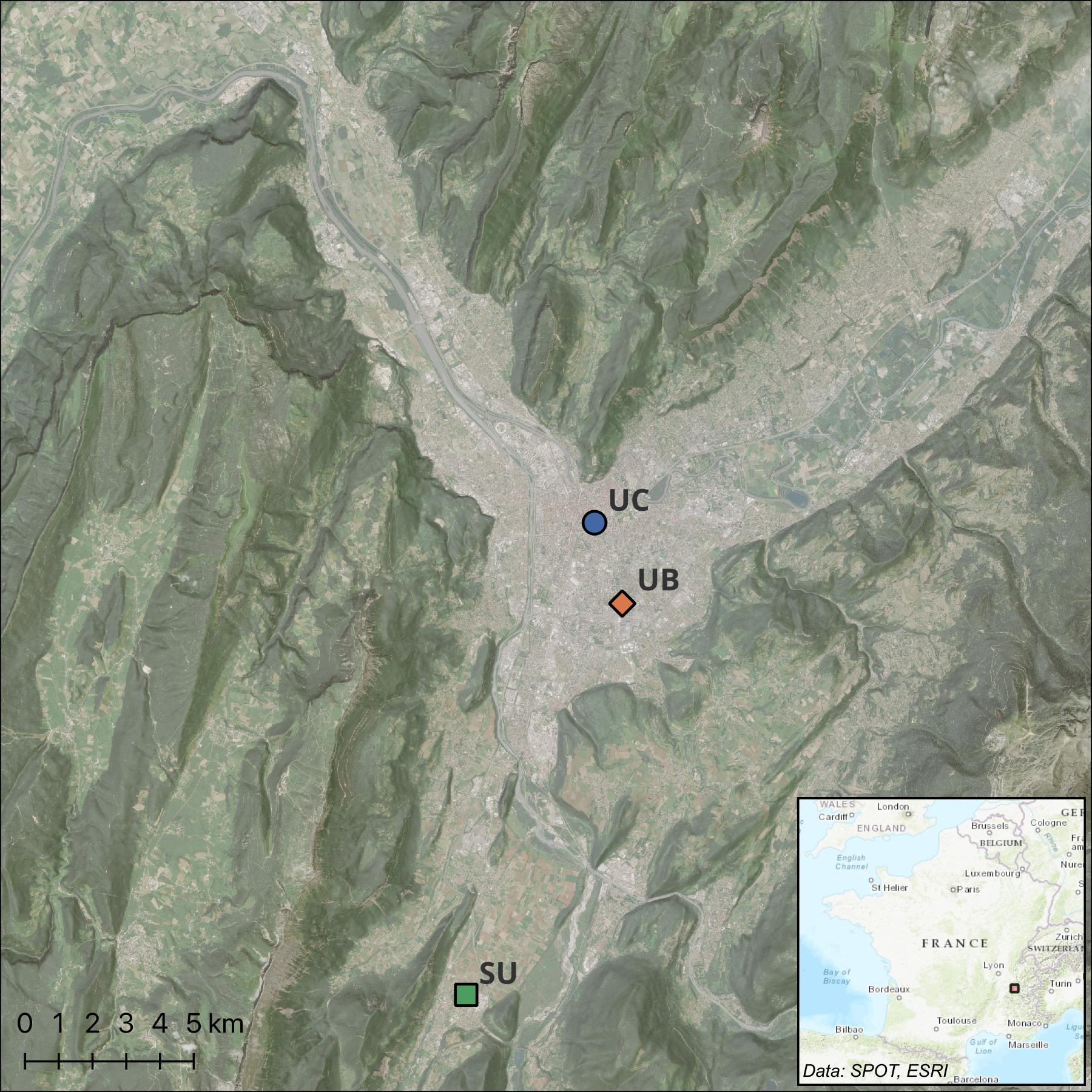}
    \caption{Study area and air quality monitoring stations. UC: urban center. UB: urban background. SU: suburban. Inset shows study area location in southeastern France.}
    \label{fig:study-area}
\end{figure}

Grenoble is a metropolitan area of about 450~000 inhabitants located in an alpine valley in southeastern France (45.2$^{\circ}$N, 5.7$^{\circ}$E) (Figure \ref{fig:study-area}). The city extends over the flat valley floor at an elevation of about 215~m above mean sea level, while the surrounding mountains rise to almost 3000~m. The mountains have a strong influence on the local meteorology, restricting airflow and contributing to atmospheric temperature inversions that trap pollutants, particularly in winter. This can produce high PM mass concentrations that extend throughout the valley, including suburban and rural areas. Biomass burning (for home heating) and sulfate- and nitrate-rich sources (related to fuel combustion) are the main contributors to PM mass, accounting on average for 68\% of PM\textsubscript{10} mass, while biomass burning and vehicle traffic are the main contributors to OP, accounting on average for 76\% of OP\textsubscript{AA} and 56\% of OP\textsubscript{DTT} \citep{Borlaza2021} (see section \ref{data-pm-op} for definitions of OP\textsubscript{AA} and OP\textsubscript{DTT}).

\subsection{Particulate matter and oxidative potential}
\label{data-pm-op}

We obtained OP from 24-hour PM\textsubscript{10} samples collected at three air quality monitoring stations on an urban to rural gradient: an urban centre station (UC, 212~m elevation), an urban background station (UB, 214~m elevation, 2.5~km south of UC), and a suburban background station (SU, 310~m elevation, 15~km south of UC) (Figure \ref{fig:study-area}). Samples at UC and SU were collected every 3 days during the periods 28 February 2017 through 11 March 2018 and 30 June 2020 through 11 July 2021. Sampling at UB was conducted every 3 days from 1 January 2017 through 31 December 2021, with daily sampling throughout the year 2019. All samples were collected using high-volume samplers (Digitel DA80, 30~m\textsuperscript{3}/hour) onto 150~mm diameter pure quartz fibre filters (Tissu quartz PALL QAT-UP 2500) following EN 12341:2014 procedures.

The OP of the PM\textsubscript{10} samples was assessed by two acellular assays. The OP\textsubscript{AA} assay measures the consumption of ascorbic acid (AA), a cellular antioxidant, and the OP\textsubscript{DTT} assay measures the consumption of dithiothreitol (DTT), a chemical surrogate that mimics interactions with biological reducing agents such as adenine dinucleotide (NADH) and nicotinamide adenine dinucleotide phosphate (NADPH). The two assays are complimentary, as the lungs contain multiple antioxidants that different in sensitivity to various chemical species. In Grenoble, OP\textsubscript{AA} is most associated with elemental carbon (related to combustion), polycyclic aromatic hydrocarbons (related to vehicle traffic), monosaccharides and potassium (related to biomass burning), organic carbon (natural and anthropogenic sources), and Sn and Cu (related to vehicle traffic). OP\textsubscript{DTT} is most associated with organic carbon (natural and anthropogenic sources), elemental carbon (related to combustion) and metals (e.g. Cu, Fe, K, Sn; related to vehicle traffic) \citep{Calas2019}.

The OP assays were performed according to the protocol developed by \citet{Calas2017, Calas2018}. Briefly, PM\textsubscript{10} was extracted from the filters into simulated lung fluid (Gamble's solution with dipalmitoylphosphatidylcholine) at an iso-mass concentration of 25~µg/mL and reacted with AA or DTT in a UV-transparent 96-well plate (CELLSTAR, Greiner-Bio). AA and DTT consumption were measured by monitoring absorbance at 265 or 412~nm for 30 minutes with a microplate reader (TECAN spectrophotometer Infinite M200 Pro). All samples were analysed in triplicate with OP recorded as the average of the three replicates. The coefficient of variation was \textless 10\% for each triplicate and \textless3\% across positive control tests. Intrinsic OP (nmol~min\textsuperscript{-1}~µg\textsuperscript{-1}) was scaled by each sample's PM\textsubscript{10} mass concentration (µg~m\textsuperscript{-3}) to give units of nmol~min\textsuperscript{-1}~m\textsuperscript{-3}, also known as oxidative burden.

\begin{figure}[t]
    \includegraphics[width=1\linewidth]{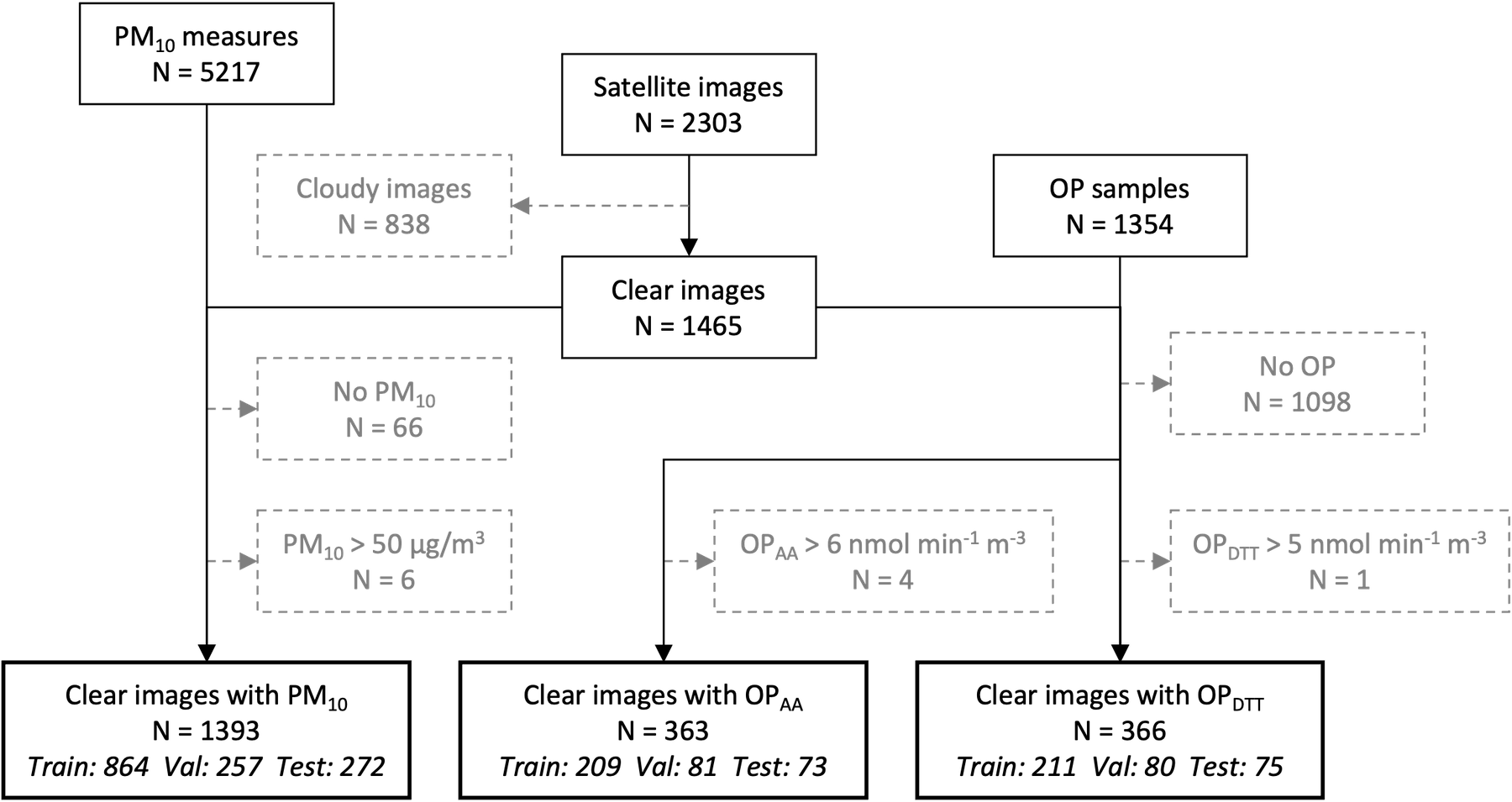}
    \caption{Number of satellite images and air quality observations}
    \label{fig:data-flow}
\end{figure}

In addition to the OP measures, we obtained daily 24-hour mean PM\textsubscript{10} mass concentration at each station from 1 January 2017 through 31 December 2021 from continuous monitoring by tapered element oscillating microbalances equipped with filter dynamics measurement systems (TEOM-FDMS). In total, we obtained 5217 measures of 24-hour mean PM\textsubscript{10} concentration and 1354 measures of 24-hour mean OP\textsubscript{AA} and OP\textsubscript{DTT}. To avoid undue influence by occasional outliers, we excluded measures where PM\textsubscript{10} \textgreater 50~µg/m\textsuperscript{3} (n = 6), OP\textsubscript{AA} \textgreater 6.0~nmol~min\textsuperscript{-1}~m\textsuperscript{-3} (n = 4), or OP\textsubscript{DTT} \textgreater 5.0~nmol~min\textsuperscript{-1}~m\textsuperscript{-3} (n = 1) (Figure \ref{sup:fig:aq-dist}). Figure \ref{fig:data-flow} shows the number of samples used for analyses.

\subsection{Meteorological data}
\label{data-met}

Meteorological conditions such as temperature, wind speed, precipitation, and mixing height directly affect PM mass concentration and composition. We obtained hourly 2-m air temperature, relative humidity, surface pressure, and wind U- and V- components at approximately 10~km spatial resolution from the ERA5-Land reanalysis \citep{Munoz-Sabater2021}. We also obtained hourly planetary boundary layer height at approximately 30~km spatial resolution from the ERA5 reanalysis \citep{Hersbach2020}. We bilinearly interpolated the data to the location of the three air quality monitoring stations and aggregated the hourly data to daily 24-hour means, yielding six meteorological variables for each station and day.

\subsection{Satellite data}
\label{data-images}

We use data from the PS2 instrument of the PlanetScope satellite constellation \citep{PlanetLabsPBC}. The PS2 instrument has four bands with a spatial resolution of approximately 3~m/pixel and the constellation of hundreds of satellites provides global coverage with daily revisits. We obtained scenes for 1 January 2017 through 31 December 2021 and clipped them to a 1~km\textsuperscript{2} square (334$\times$334~pixels) centred on each air quality monitoring station, merging adjacent scenes from the same satellite overpass as needed to ensure at least 90\% cover of the area around each station. On days with multiple satellite overpasses, we kept only the scene with the lowest cloud cover or from the overpass latest in the day.

We downloaded two image types for each scene. Visual images (RGB) have red, green, and blue bands and are colour-corrected and sharpened to facilitate interpretation by the human eye. Analytic images (TOAR) have blue, green, red, and near-infrared bands and were scaled to top-of-atmosphere reflectance before downloading. The RGB images are similar to the three-channel images that are widely used in computer vision tasks, while the TOAR images' extra near-infrared band might contain additional information related to air quality.

\begin{figure}[h]
    \includegraphics[width=1\linewidth]{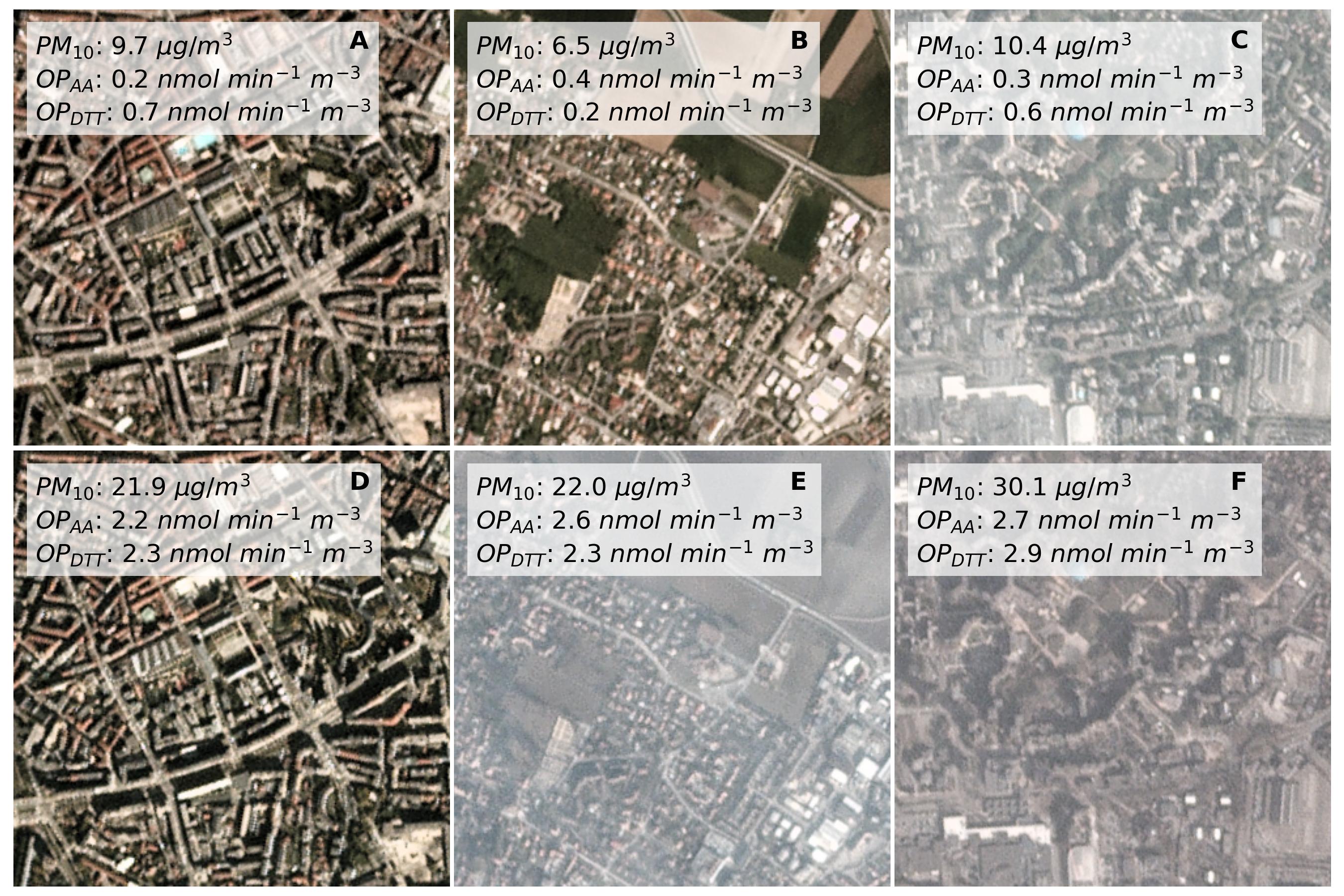}
    \caption{Example satellite images illustrating that visual haziness is not consistently correlated with air quality. Top row: days with low low PM\textsubscript{10}, OP\textsubscript{AA}, and OP\textsubscript{DTT} at stations UC (A), SU (B), and UB (C). Bottom row (D-E): days with high low PM\textsubscript{10}, OP\textsubscript{AA}, and OP\textsubscript{DTT}.}
    \label{fig:images}
\end{figure}

The scene cloud cover metadata was very inaccurate for our small areas of interest, so we filtered out cloudy images based on the values of the TOAR green band and visual inspection (details in \ref{sup:filter}). We associated each clear (not cloudy) image with meteorological variables, PM\textsubscript{10} mass concentration, OP\textsubscript{AA}, and OP\textsubscript{DTT}. Figure \ref{fig:data-flow} shows the number of images and corresponding air quality observations and Figure \ref{fig:images} shows six example images.

\section{Methods}

\subsection{Network architecture}

\begin{figure}[t]
    \includegraphics[width=1\linewidth]{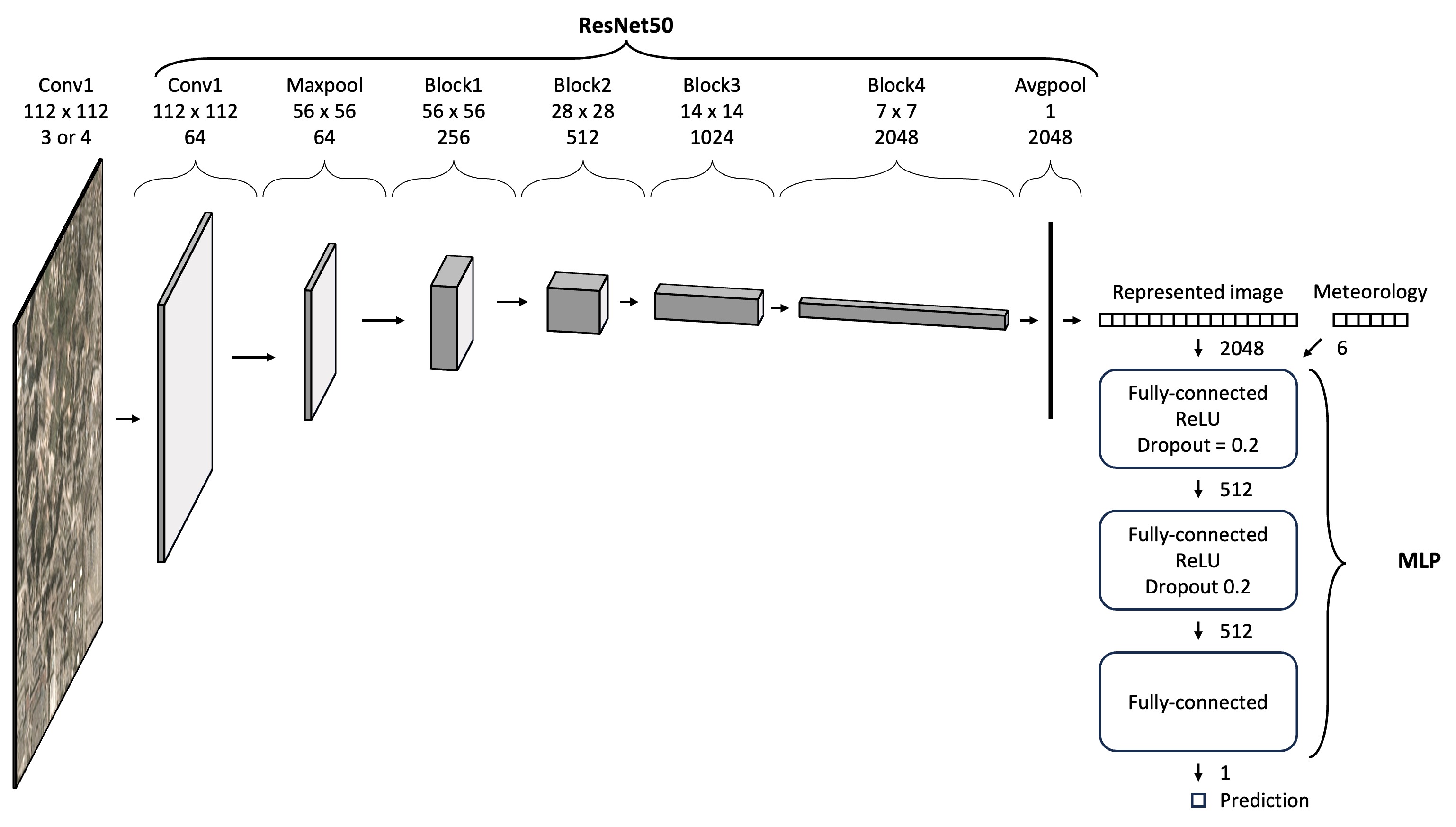}
    \caption{Network architecture. Inputs and outputs are labelled by size (Height x Width x Channels).}
    \label{fig:architecture}
\end{figure}

Our network architecture is based on previous work by \citet{Zheng2020, Zheng2021} and \citet{Jiang2022}, who used a similar architecture to estimate 24-hour mean PM\textsubscript{2.5} mass concentration from PlanetScope images in Beijing and Delhi. The architecture consists of a ResNet50 CNN \citep{He2016} whose fully-connected output layer has been replaced by a 3-layer multilayer perceptron (MLP) (Figure \ref{fig:architecture}). To estimate air quality, a satellite image of dimension 334$\times$334~pixels with three (RGB images) or four (TOAR images) channels is fed into the CNN, which extracts a 2048-feature representation of the image and passes it to the MLP. The MLP’s fully-connected input layer and hidden layer have output dimension 512, ReLU activation, and a dropout rate of 0.2. The fully-connected output layer returns a single value, the estimated PM\textsubscript{10} mass concentration, OP\textsubscript{AA}, or OP\textsubscript{DTT}. For models that use meteorology, the six meteorological variables (section \ref{data-met}) are concatenated to the CNN-extracted representation of the image (yielding a 2054-feature representation) before passing to the MLP.

\subsection{Experimental setup}

We perform a series of experiments to assess the potential for estimating PM\textsubscript{10} mass concentration and OP from satellite images and meteorological variables. We perform each experiment separately for each of PM\textsubscript{10} mass concentration, OP\textsubscript{AA}, and OP\textsubscript{DTT}. All experiments use the same randomly selected 60\% train set, 20\% validation set, and 20\% test set, with all images from the same day assigned to the same set. Figure \ref{fig:data-flow} shows the number of training, validation, and test samples. To facilitate model convergence, we normalize all images by subtracting the per-channel mean and dividing by the per-channel standard deviation of the training set images.

We evaluate the experiments based on the coefficient of determination (R\textsuperscript{2}), root mean squared error (RMSE), and normalized mean absolute error (NMAE) as defined in Equations \ref{eq:r2}-\ref{eq:nmae}.

\begin{equation}
\text{R}^2 = 1 - \frac{\sum (y - \hat{y})^2}{\sum (y-\bar{y})^2}
\label{eq:r2}
\end{equation}

\begin{equation}
\text{RMSE} = \sqrt{\frac{\sum (y - \hat{y})^2}{n}}
\label{eq:rmse}
\end{equation}

\begin{equation}
\text{NMAE} = \frac{\sum |y - \hat{y}|}{n \cdot \bar{y}}
\label{eq:nmae}
\end{equation}

where $y$ are the observed values, $\hat{y}$ are the model's estimates, $\bar{y}$ is the mean observed value, and $n$ is the number of observations. We estimate the uncertainty of the scoring metrics by bootstrapping: we compute the variation of the metrics across 1000 resamples of the test set drawn randomly with replacement. We run all experiments on a MacBook Air M1 with 16~GB RAM using \texttt{python 3.11}, \texttt{pytorch 2.1.2}, \texttt{pytorch-lightning 2.1.3}, and \texttt{torchvision 0.16.2}.

\subsection{Baseline model}
\label{exp-baseline}

We first develop a \textbf{baseline model} that uses only the 6 meteorological variables (section \ref{data-met}), which are fed directly into the MLP. This model evaluates the capacity to estimate air quality using only data that are widely available and commonly used to estimate PM mass concentration. To allow direct comparisons with models that use satellite images, we train and evaluate the baseline model using only data from the station days for which clear satellite images are available (Figure \ref{fig:data-flow}). We train the MLP to minimize mean squared error loss on mini batches of size 32, dropping the last batch if it contains fewer than 32 samples. We train for 150 epochs using the Adam optimizer \citep{Kingma2017} with a fixed learning rate of 0.0005.

\subsection{Random features}
\label{exp-random}

Our second model uses random features of the satellite images. For the \textbf{random model}, we initialize the ResNet50 CNN with random weights and use it to extract features from the satellite images. We then concatenate the 6 meteorological variables (section \ref{data-met}) to the image features and feed the resulting 2054-element vectors into the MLP. We train the MLP using the same approach as for the baseline model (section \ref{exp-baseline}) but stop training as soon as the validation loss fails to improve after 25 epochs.

\subsection{Transfer learning}
\label{exp-transfer}

Transfer learning uses features extracted by a deep CNN that has been trained on a large image dataset as predictors in a new learning task. This approach has been shown to give good performance on a wide variety of vision tasks thanks to the generic nature of the extracted features \citep{Razavian2014}, and was recently applied by \citet{Zheng2020} and \citet{Jiang2022} to estimate 24-hour mean PM\textsubscript{2.5} mass concentration based on PlanetScope RGB images in Beijing and Delhi. For the \textbf{transfer model}, we initialize the ResNet50 CNN with weights learned on the ImageNet 1K classification task \citep{Deng2010}. We use the CNN to extract features of the images, concatenate the meteorological variables, and train the MLP using the same approach as for the random model (section \ref{exp-random}): mean squared error loss, mini-batches of size 32, Adam optimizer with fixed learning rate of 0.0005, training for up to 150 epochs with early stopping if the validation loss fails to improve after 25 epochs.

\subsection{Fine-tuning}
\label{exp-tune}

Since ImageNet does not include any satellite images, ImageNet-derived features may be less than optimal for our air quality estimation task. We address this by fine-tuning the features to adapt them to our data and learning task. For the \textbf{fine-tuning model}, we initialize the ResNet50 CNN with weights from ImageNet as for the transfer model and freeze all but the final \texttt{Block4} and \texttt{Avgpool} layers (Figure \ref{fig:architecture}). We then train those layers simultaneously with the MLP using the same approach as for the transfer model (section \ref{exp-transfer}).

\subsection{Contrastive learning}
\label{exp-simsiam}

Fine-tuning can adapt generic features to a specific dataset or task, but the fine-tuned features may still be sub-optimal since only the shallowest layers of the CNN are retrained. Supervised fine-tuning also requires a large and diverse labelled training dataset. An alternative approach is to use a form of self-supervised learning, such as contrastive learning, which trains a CNN to extract features that capture similarities and differences between a large dataset of unlabelled images. Recently, \citet{Jiang2022} found that contrastive learning improved estimates of 24-hour mean PM\textsubscript{2.5} mass concentration compared to transfer learning, particularly when using a small number of air quality monitoring stations.

SimSiam is a simple contrastive learning framework that has shown good performance on the ImageNet classification task \citep{Chen2021} and for estimating PM\textsubscript{2.5} in Beijing and Delhi \citep{Jiang2022}. Briefly, SimSiam uses a deep CNN (here ResNet50) whose output layer is replaced by an MLP. The modified CNN outputs an intermediate representation for each of two augmented views of a single image. The intermediate representations are passed to a second MLP that outputs a final representation for each. The entire framework is trained to maximize the cosine similarity between the intermediate representation of one view and the final representation of the other view across a large set of images. A stop gradient is applied to each intermediate representation to prevent the CNN from collapsing to a state where it outputs the same representation for every image. After training, the MLPs are discarded, and the trained CNN can be applied to supervised learning tasks.

For the \textbf{SimSiam model} model, we initialize the ResNet50 CNN with random weights and use the SimSiam framework to train it on \textit{all} clear train set images, including images with no corresponding air quality observation (n = 906). We generate augmented views by taking a randomly-located crop of 20\% to 100\% of the image, resizing to 96$\times$96 pixels (to limit the computational burden), and flipping horizontally with probability 0.5. We do not apply any colour or blurring augmentation, as we hypothesize that changes in colour or sharpness may be related to PM mass concentration or OP. We train on mini batches of 32 images using stochastic gradient descent with momentum of 0.9 and weight decay 0.0001. We use cosine annealing to decay the learning rate from 0.005 to 0 over the course of 100 epochs. We then freeze the ResNet50 CNN and proceed to train the MLP using the same approach as for the transfer learning model (section \ref{exp-transfer}).

To evaluate the potential of contrastive features learned from a larger dataset of satellite images, we also develop \textbf{SimSiam BJ} and \textbf{SimSiam DL} models. These models transfer contrastive features that were learned from PlanetScope images of Beijing and Delhi, respectively \citep{Jiang2022}. Since Beijing and Delhi are much larger, more polluted, and have different emissions sources, topography, and meteorology compared to Grenoble, we adapt the features using the same approach as for the fine-tuning model (section \ref{exp-tune}). Specifically, we initialize the ResNet50 CNN with pretrained weights learned in Beijing or Delhi, freeze all but the final \texttt{Block4} and \texttt{Avgpool} layers, and train those layers simultaneously with the MLP.

\subsection{Further experiments}

Our main models use RGB images and meteorological variables. To better evaluate the images’ contribution to the air quality estimates, we also train a version of each model that uses only image features. To evaluate whether the near-infrared band contains additional information related to air quality, we repeat the transfer, fine-tuning, and SimSiam experiments using four-channel TOAR images rather than three-channel RGB images. For the TOAR experiments, we modify the ResNet50’s Conv1 layer (Figure \ref{fig:architecture}) to accept four-channel inputs. We initialize the added weights with random values, keeping the original values for all other weights.

As a final experiment, we evaluate whether the \textbf{transfer} and \textbf{fine-tuning} models fully leverage the multi-modal data, which has very different dimensions (2048 image features vs. 6 meteorological variables). We do this by refitting the models that use a high-dimensional (n = 1976) sparse binary representation of the six meteorological variables, as suggested by \citet{Zheng2021}. We use \texttt{sklearn}’s \texttt{RandomTreesEmbedding} to create an unsupervised extremely randomized forest \citep{Geurts2006} with 256 trees of max depth 3 (up to 8 leaves per tree). The forest is trained to separate observed meteorology datapoints from synthetic datapoints sampled from the joint distribution of the meteorological variables. The forest then classifies the observed datapoints and one-hot encodes them with the index of the leaf nodes into which they fall. This results in a vector for each datapoint that has up to 2048 values (256$\times$2\textsuperscript{3}) of which 256 are 1 (one per tree) and the rest are 0.

\section{Results and discussion}

\subsection{Air quality}

Table \ref{tab:aq-dist} lists the mean and standard deviation of observed PM\textsubscript{10} mass concentration, OP\textsubscript{AA}, and OP\textsubscript{DTT}. Mean PM\textsubscript{10} mass concentration, OP\textsubscript{AA}, and OP\textsubscript{DTT} were highest at the urban centre station (UC), slightly lower at the urban background station (UB), and lowest at the suburban station (SU). The air quality measures followed approximately log-normal distributions, with PM\textsubscript{10} ranging from 1.5 to 84.3~µg/m\textsuperscript{3}, OP\textsubscript{AA} from 0.005 to 9.5~nmol~min\textsuperscript{-1}~m\textsuperscript{-3}, and OP\textsubscript{DTT} from 0.003 to 9.86~nmol~min\textsuperscript{-1}~m\textsuperscript{-3} (Figure \ref{sup:fig:aq-dist}).

\begin{table}[!b]
    \begin{threeparttable}
        \caption{Descriptive statistics of PM\textsubscript{10}, OP\textsubscript{AA}, and OP\textsubscript{DTT}.}
        \label{tab:aq-dist}
        \begin{tabular}{lrccrccrcc} 
            \toprule
            & \multicolumn{3}{c}{PM\textsubscript{10}}& \multicolumn{3}{c}{OP\textsubscript{AA}} & \multicolumn{3}{c}{OP\textsubscript{DTT}}\\
            \cmidrule(lr){2-4}
            \cmidrule(lr){5-7}
            \cmidrule(lr){8-10}
            Station & N & Mean & SD & N & Mean & SD & N & Mean & SD \\
            \midrule
            UC& 1683& 18.3& 9.2& 246& 1.51& 1.34& 246& 1.61& 1.07\\ 
            UB& 1755& 17.1& 9.1& 856& 1.40& 1.26& 856& 1.40& 1.04\\ 
            SU& 1779& 13.9& 8.5& 252& 1.27& 1.52& 252& 1.12& 0.80\\ 
            \textit{Total}& \textit{5217}& \textit{16.4}& \textit{9.1}& \textit{1354}& \textit{1.39}& \textit{1.33}& \textit{1354}& \textit{1.38}& \textit{1.02}\\ 
            \bottomrule
        \end{tabular}
        \begin{tablenotes}
            \item N = number of measures; SD = standard deviation. Mean and SD units are µg/m\textsuperscript{3} for PM\textsubscript{10} and nmol~min\textsuperscript{-1}~m\textsuperscript{-3} for OP\textsubscript{AA} and OP\textsubscript{DTT}.
        \end{tablenotes}
    \end{threeparttable}
\end{table}

\begin{figure}[h]
    \includegraphics[width=1\linewidth]{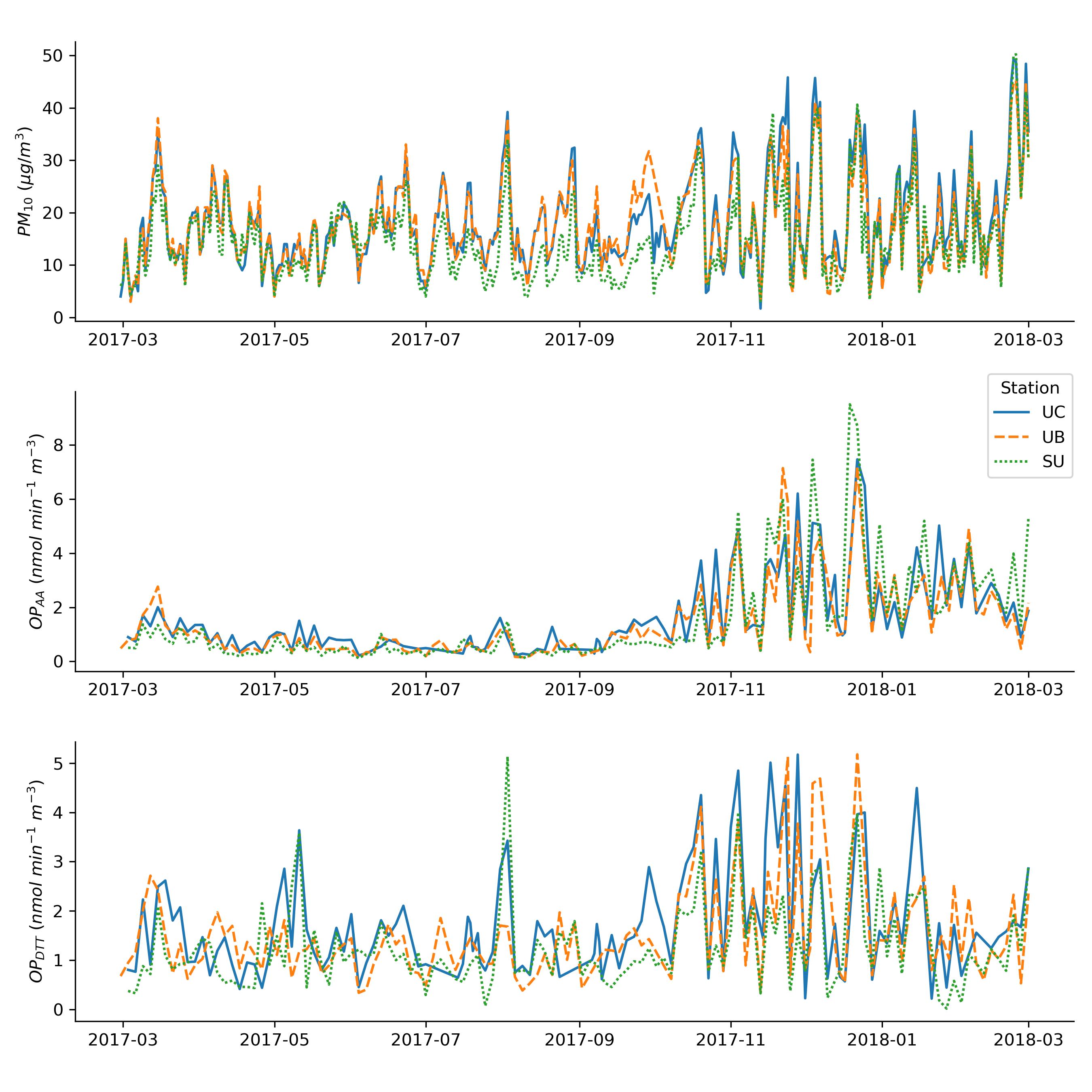}
    \caption{Observed PM\textsubscript{10} (A), OP\textsubscript{AA} (B), and OP\textsubscript{DTT} (C) from 2017-02-08 to 2018-03-11}
    \label{fig:timeseries}
\end{figure}

Figure \ref{fig:timeseries} shows one year of air quality observations. OP\textsubscript{AA} and OP\textsubscript{DTT} were higher and more variable during the winter, although OP\textsubscript{DTT} also showed occasional peaks in spring and summer. PM\textsubscript{10} mass concentration was strongly correlated between stations (Spearman $\rho \geq 0.88$) while OP\textsubscript{AA} and OP\textsubscript{DTT} were less correlated between stations ($\rho \geq 0.81$ for OP\textsubscript{AA}; $\rho \geq 0.58$ for  OP\textsubscript{DTT}). Correlations were strongest between UC and UB and weakest between UC and SU (Table \ref{sup:tab:station-cor}). PM\textsubscript{10} was more correlated with OP\textsubscript{DTT} ($\rho$ = 0.79) than with OP\textsubscript{AA} ($\rho$ = 0.59) while OP\textsubscript{AA} and OP\textsubscript{DTT} were moderately correlated ($\rho$ = 0.65). 

\subsection{OP\textsubscript{AA} models}

\subsubsection{OP\textsubscript{AA}: baseline performance}

The baseline OP\textsubscript{AA} model (which uses only meteorology) captures more than half of the variation in OP\textsubscript{AA}, with test set R\textsuperscript{2} of 0.60, RMSE of 0.71~nmol~min\textsuperscript{-1}~m\textsuperscript{-3}, and NMAE of 35\% (Table \ref{tab:scores}). This is consistent with the strong seasonality of OP\textsubscript{AA} shown in Figure \ref{fig:timeseries} and the fact that OP\textsubscript{AA} is moderately correlated with both air temperature (Spearman $\rho$ = -0.55) and boundary layer height ($\rho$ = -0.74). The random model (which uses random features of RGB images) is slightly less accurate than the baseline model when using both image features and meteorology (Table \ref{tab:scores}), and much less accurate when using only image features (R\textsuperscript{2} = 0.23, RMSE = 0.98~nmol~min\textsuperscript{-1}~m\textsuperscript{-3}, NMAE = 48\%). These results indicate that random image features are only weakly related to OP\textsubscript{AA}. The baseline and random models show some overfitting, with better performance on the training and validation sets compared to the test set (Figure \ref{fig:scatter-aa-met}; Figure \ref{sup:fig:scatter-aa}). This may in part be due to the fact that the test set OP\textsubscript{AA} values skew higher and are more variable than the training and validation set values.

\begin{table}[!htb]
    \begin{threeparttable}
        \caption{Test set performance of the PM\textsubscript{10}, OP\textsubscript{AA}, and OP\textsubscript{DTT} models.}
        \label{tab:scores}
        \small
        \begin{tabular}{llrrrrrrrrr}
            \toprule
            \multicolumn{2}{c}{} & \multicolumn{3}{c}{OP\textsubscript{AA}} & \multicolumn{3}{c}{OP\textsubscript{DTT}} & \multicolumn{3}{c}{PM\textsubscript{10}}\\
            \cmidrule(lr){3-5}
            \cmidrule(lr){6-8}
            \cmidrule(lr){9-11}
            Model & Features & R\textsuperscript{2} & RMSE & NMAE & R\textsuperscript{2} & RMSE & NMAE & R\textsuperscript{2} & RMSE & NMAE \\
            \midrule
            Baseline    & M   & 0.60 & 0.71 & 35\% & 0.30 & 0.78 & 37\% & 0.33 & 5.9 & 27\% \\
            Random      & I+M & 0.55 & 0.75 & 36\% & 0.29 & 0.79 & 37\% & 0.43 & 5.5 & 25\% \\
            \addlinespace
            Transfer    & I+M & 0.54 & 0.76 & 33\% & 0.39 & 0.73 & 36\% & 0.31 & 6.0 & 27\% \\
            Fine-tuning & I+M & 0.60 & 0.71 & 34\% & 0.40 & 0.73 & 35\% & 0.37 & 5.8 & 25\% \\
            Transfer    & I+H & 0.62 & 0.69 & 31\% & 0.48 & 0.67 & 32\% & 0.45 & 5.4 & 24\% \\
            Fine-tuning & I+H & 0.63 & 0.68 & 31\% & 0.48 & 0.67 & 33\% & 0.44 & 5.4 & 22\% \\
            \addlinespace
            SimSiam     & I+M & 0.44 & 0.84 & 40\% & 0.24 & 0.82 & 40\% & 0.29 & 6.2 & 28\% \\
            SimSiam BL  & I+M & 0.15 & 1.03 & 50\% & 0.10 & 0.89 & 42\% & 0.12 & 6.8 & 30\% \\
            SimSiam DL  & I+M & 0.42 & 0.86 & 43\% & 0.31 & 0.78 & 38\% & 0.16 & 6.7 & 30\% \\
            \addlinespace
            Random      & I   & 0.23 & 0.98 & 48\% & 0.18 & 0.85 & 38\% & 0.16 & 6.7 & 30\% \\
            Transfer    & I   & 0.48 & 0.81 & 36\% & 0.35 & 0.76 & 37\% & 0.23 & 6.4 & 28\% \\
            Fine-tuning & I   & 0.49 & 0.80 & 38\% & 0.36 & 0.75 & 37\% & 0.22 & 6.4 & 26\% \\
            SimSiam     & I   & 0.14 & 1.04 & 48\% & 0.15 & 0.87 & 40\% & 0.12 & 6.8 & 31\% \\
            SimSiam BL  & I   & 0.17 & 1.02 & 50\% & 0.01 & 0.94 & 45\% & 0.09 & 6.9 & 32\% \\
            SimSiam DL  & I   & 0.45 & 0.83 & 41\% & 0.23 & 0.82 & 40\% & 0.19 & 6.5 & 28\% \\
            \bottomrule
        \end{tabular}
        \begin{tablenotes}
            \item M = meteorological variables (n = 6). I+M = RGB image features (n = 2048) and meteorological variables. I+H = RGB image features and high-dimensional representation of meteorology (n = 1976). I = RGB image features. RMSE units are µg/m\textsuperscript{3} for PM\textsubscript{10} and nmol~min\textsuperscript{-1}~m\textsuperscript{-3} for OP\textsubscript{AA} and OP\textsubscript{DTT}.
        \end{tablenotes}
    \end{threeparttable}
\end{table}

\subsubsection{OP\textsubscript{AA}: transfer learning and fine-tuning}

The transfer and fine-tuning OP\textsubscript{AA} models (which use ImageNet-derived features of RGB images) perform similarly to the baseline model when using the six meteorological variables, and slightly better when using a high-dimensional representation of meteorology (Table \ref{tab:scores}). This suggests that the images contain little new information related to OP\textsubscript{AA} beyond the information present in the meteorological variables. When using only image features, the transfer model captures about half of the variation in OP\textsubscript{AA}, with test set R\textsuperscript{2} of 0.48, RMSE of 0.81~nmol~min\textsuperscript{-1}~m\textsuperscript{-3}, and NMAE of 36\%. This suggests that about 80\% of the OP\textsubscript{AA}-related information in the meteorological variables can be derived from the images. Using four-channel TOAR images slightly increases the accuracy of the fine-tuning model that uses only image features (R\textsuperscript{2} = 0.53, RMSE = 0.77~nmol~min\textsuperscript{-1}~m\textsuperscript{-3}, NMAE = 37\%) (Table \ref{sup:tab:scores-toar}). The near-infrared band is important for vegetation phenology tracking, so including it may help the model identify OP\textsubscript{AA}’s strong seasonal trend (Figure \ref{fig:timeseries}).

\begin{figure}
    \includegraphics[width=0.9\linewidth]{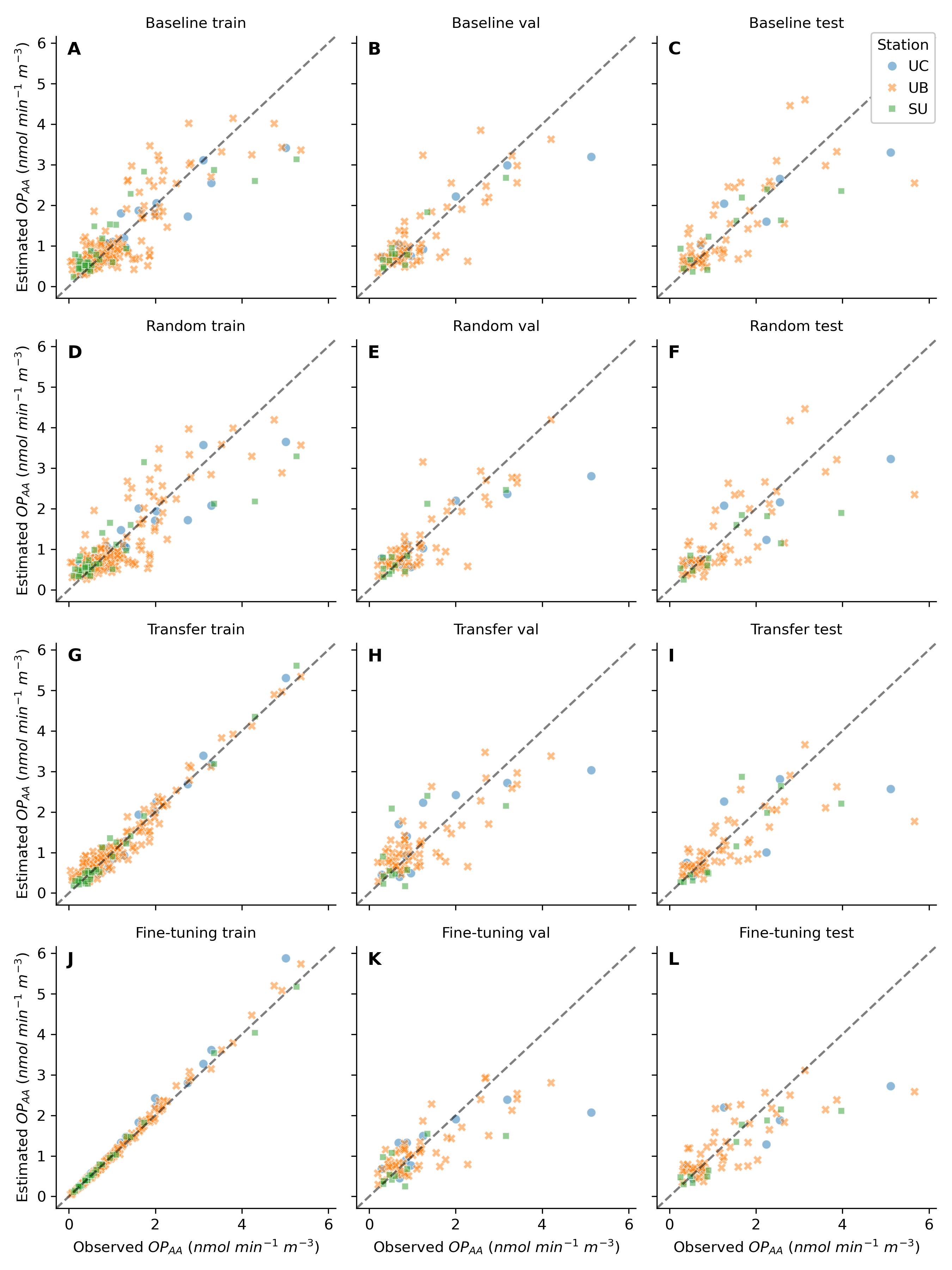}
    \caption{Observed and estimated OP\textsubscript{AA} for the training (left column), validation (middle column), and test (right column) sets. Top row (A-C): baseline model (meteorology only). Second row (D-F): random model (meteorology and random features of RGB images). Third row (G-I): transfer model (meteorology and ImageNet-derived features of RGB images). Bottom row (J-L): fine-tuning model (meteorology and fine-tuned ImageNet-derived features of RGB images).}
    \label{fig:scatter-aa-met}
\end{figure}

The transfer and fine-tuning models give much more accurate estimates for the training set than the validation and test sets (Figure \ref{fig:scatter-aa-met}; Figure \ref{sup:fig:scatter-aa}, Table \ref{sup:tab:scores-aa}). This suggests that ImageNet-derived features can accurately distinguish between images of different stations and days, but the features that discriminate between images are not always related to OP\textsubscript{AA}.

\subsubsection{OP\textsubscript{AA}: contrastive learning}

The SimSiam OP\textsubscript{AA} model (which uses contrastive features of RGB images selected by SimSiam) is less accurate than the models that use ImageNet-derived features (Table \ref{tab:scores}). When using only image features, the SimSiam model performs similarly to the random model, with test set R\textsuperscript{2} of 0.14, RMSE of 1.0~nmol~min\textsuperscript{-1}~m\textsuperscript{-3} and NMAE of 48\%. Although using four-channel TOAR images improves accuracy (R\textsuperscript{2} = 0.35, RMSE = 0.90~nmol~min\textsuperscript{-1}~m\textsuperscript{-3}, NMAE = 44\%) (Table \ref{sup:tab:scores-toar}), it fails to match the performance of the transfer and fine-tuning models.

The poor performance of the SimSiam model may be due to the limited number of images used for contrastive learning (n = 906). Repeating the contrastive learning using all available images (n = 1465 after including validation and test images) does not substantially improve the results. However, the SimSiam DL model (which fine-tunes contrastive features learned from 31 475 PlanetScope images in Delhi) produces more accurate estimates (Table \ref{tab:scores}). When using only image features, the SimSiam DL model performs almost as well as the transfer and fine-tuning models, with test set R\textsuperscript{2} of 0.45, RMSE of 0.83~nmol~min\textsuperscript{-1}~m\textsuperscript{-3}, and NMAE of 41\%. This could point to similarities in urban form, emissions sources, or PM components between Delhi and Grenoble, or it may be that the Delhi dataset is diverse enough for contrastive learning to derive generic features that are related to both PM mass concentration in Delhi and OP\textsubscript{AA} in Grenoble. In contrast, the SimSiam BJ model (which fine-tunes contrastive features learned from 13~022 PlanetScope images in Beijing) performs similarly to the SimSiam and random models, with test set R\textsuperscript{2} of 0.17, RMSE of 1.0~nmol~min\textsuperscript{-1}~m\textsuperscript{-3}, and NMAE of 50\%. The SimSiam BJ model does perform better than the SimSiam and random models on the training set (Table \ref{sup:tab:scores-aa}). This suggests that contrastive features learned in Beijing can discriminate between images in Grenoble, but differences in urban form or other factors between the two cities mean that the features are unrelated to OP\textsubscript{AA} in Grenoble.

Supplementary Figures \ref{sup:fig:loss-aa-met} and \ref{sup:fig:loss-aa-nomet} show training and validation loss curves for the OP\textsubscript{AA} models.

\subsection{OP\textsubscript{DTT} models}

\subsubsection{OP\textsubscript{DTT}: baseline performance}

The baseline OP\textsubscript{DTT} model (which uses only meteorology) captures about one third of the variation in OP\textsubscript{DTT}, with test set R\textsuperscript{2} of 0.30, RMSE of 0.78~nmol~min\textsuperscript{-1}~m\textsuperscript{-3}, and NMAE of 37\% (Table \ref{tab:scores}). The lower accuracy compared to the baseline OP\textsubscript{AA} model is consistent with OP\textsubscript{DTT}’s weaker seasonality (Figure \ref{fig:scatter-dtt-met}) and limited correlation with meteorological variables. The random OP\textsubscript{DTT} model (which uses random features of RGB images) performs very similarly to the baseline model when using both image features and meteorology (Table \ref{tab:scores}) but performs less well when using only image features (R\textsuperscript{2} = 0.18, RMSE = 0.85~nmol~min\textsuperscript{-1}~m\textsuperscript{-3}, NMAE = 38\%). The baseline and random models show little overfitting, with similar performance on the validation and test sets (Figure \ref{fig:scatter-dtt-met}, Figure \ref{sup:fig:scatter-dtt}, Table \ref{sup:tab:scores-dtt}).

\subsubsection{OP\textsubscript{DTT}: transfer learning and fine-tuning}

\begin{figure}
    \includegraphics[width=0.9\linewidth]{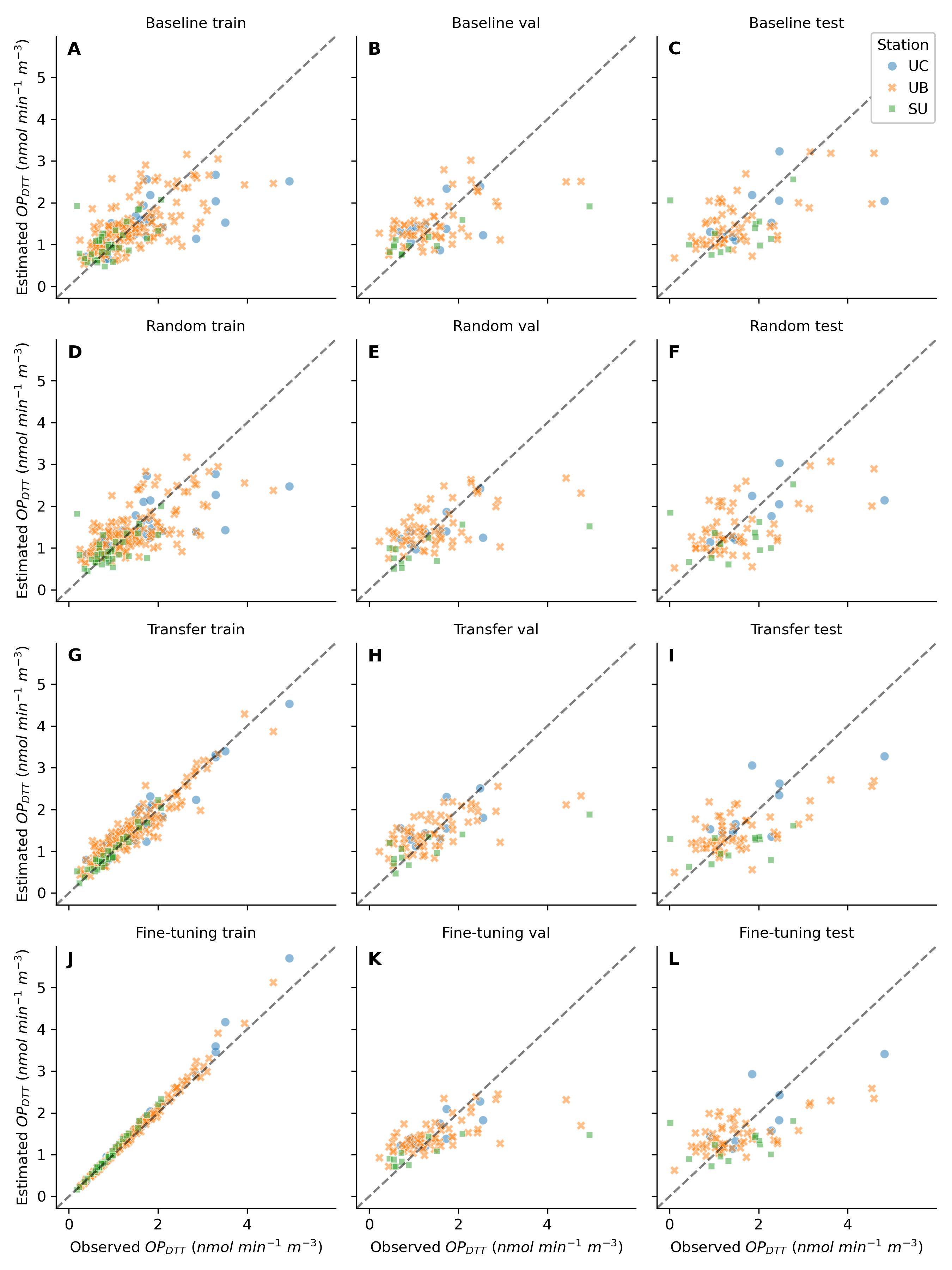}
    \caption{Observed and estimated OP\textsubscript{DTT} for the training (left column), validation (middle column), and test (right column) sets. Top row (A-C): baseline model (meteorology only); second row (D-F): random model (meteorology and random features of RGB images); third row (G-I): transfer model (meteorology and ImageNet-derived features of RGB images); bottom row (J-L): fine-tuning model (meteorology and fine-tuned ImageNet-derived features of RGB images).}
    \label{fig:scatter-dtt-met}
\end{figure}

The transfer and fine-tuning OP\textsubscript{DTT} models (which use ImageNet-derived features of RGB images) are more accurate than the baseline and random models (Table \ref{tab:scores}). When using both image features and a high-dimensional representation of meteorology, the transfer model captures about half of the variation in OP\textsubscript{DTT}, with test set R\textsuperscript{2} of 0.48, RMSE of 0.67~nmol~min\textsuperscript{-1}~m\textsuperscript{-3}, and NMAE of 32\%. When using only image features, it still captures about a third of the variation in OP\textsubscript{DTT} (R\textsuperscript{2} = 0.35, RMSE = 0.76~nmol~min\textsuperscript{-1}~m\textsuperscript{-3}, NMAE = 37\%). Using four-channel TOAR images slightly increases the fine-tuning model’s accuracy when using only image features (R\textsuperscript{2} = 0.40, RMSE = 0.72~nmol~min\textsuperscript{-1}~m\textsuperscript{-3}, NMAE = 36\%) (Table \ref{sup:tab:scores-toar}). These results suggest that images and meteorology contain complimentary information related to OP\textsubscript{DTT} and highlight the importance of using similar-dimension representations to help the model leverage the multi-modal data. Fine-tuning the ImageNet-derived features increases accuracy on the training set (Figure \ref{fig:scatter-dtt-met}, Figure \ref{sup:fig:scatter-dtt}, Table \ref{sup:tab:scores-dtt}), but barely changes performance on the test set. This suggests that fine-tuning improves the model’s ability to discriminate between images, but the improvements are mostly unrelated to OP\textsubscript{DTT}.

The lower R\textsuperscript{2} of the OP\textsubscript{DTT} transfer model compared to the OP\textsubscript{AA} model (0.35 vs 0.48 when using only image features) is due to the lower variability of OP\textsubscript{DTT}. The mean observed value of OP\textsubscript{DTT} and OP\textsubscript{AA} is very similar (1.38 vs. 1.39~nmol~min\textsuperscript{-1}~m\textsuperscript{-3}), but the standard deviation of OP\textsubscript{DTT} is about 74\% of the mean while the standard deviation of OP\textsubscript{AA} is about 96\% of the mean (Table \ref{tab:aq-dist}). Indeed, the OP\textsubscript{DTT} transfer model has similar or lower RMSE and NMAE compared to the OP\textsubscript{AA} transfer model (Table \ref{tab:scores}).

\subsubsection{OP\textsubscript{DTT}: contrastive learning}

The SimSiam OP\textsubscript{DTT} model (which uses contrastive features of RGB images selected by SimSiam) is less accurate than the models that use ImageNet-derived features (Table \ref{tab:scores}). When using only image features, the SimSiam model performs similarly to the random model, with test set R\textsuperscript{2} of 0.15, RMSE of 0.87~nmol~min\textsuperscript{-1}~m\textsuperscript{-3} and NMAE of 40\%. Although using four-channel TOAR images improves accuracy (R\textsuperscript{2} = 0.29, RMSE = 0.79~nmol~min\textsuperscript{-1}~m\textsuperscript{-3}, NMAE = 37\%) (Table \ref{sup:tab:scores-toar}), it fails to match the performance of the transfer and fine-tuning models.

As with OP\textsubscript{AA}, repeating the contrastive learning on all images (n = 1465) does not substantially change performance. Fine-tuning contrastive features learned from a much larger image dataset (31 475 RGB images in Delhi) increases accuracy, but the SimSiam DL model still does not match the performance of the transfer and fine-tuning models (Table \ref{tab:scores}). The SimSiam BJ model performs very poorly on the test set: when using only image features, it has test set R\textsuperscript{2} of -0.01, indicating that its estimates are no closer to the observed values than the mean observed OP\textsubscript{DTT} (Table \ref{tab:scores}).

Supplementary Figures \ref{sup:fig:loss-dtt-met} and \ref{sup:fig:loss-dtt-nomet} show training and validation loss curves for the OP\textsubscript{DTT} models.

\subsection{PM\textsubscript{10} models}

\subsubsection{PM\textsubscript{10}: baseline performance}

The baseline PM\textsubscript{10} model (which uses only meteorology) captures about one third of the variation in PM\textsubscript{10} mass concentration, with test set R\textsuperscript{2} of 0.33, RMSE of 5.9 µg/m3, and NMAE of 27\% (Table \ref{tab:scores}). This is consistent with PM\textsubscript{10}’s weak seasonality (Figure \ref{fig:scatter-pm-met}) and limited correlation with meteorological variables. The model’s low NMAE despite fairly low R\textsuperscript{2} is due to the low variability of PM\textsubscript{10}, whose standard deviation is about 55\% of the mean (Table \ref{tab:aq-dist}). The random model (which uses random features of RGB images) is slightly more accurate than the baseline model, with test set R\textsuperscript{2} of 0.43, RMSE of 5.5 µg/m3, and NMAE of 25\% when using image features and meteorology. But the random model is less accurate than the baseline model on the training and validation sets (Table \ref{sup:tab:scores-pm}) and is less accurate on the test set when using only image features (R\textsuperscript{2} = 0.16, RMSE = 6.7 µg/m3, NMAE = 30\%).

\subsubsection{PM\textsubscript{10}: transfer learning and fine-tuning}

The transfer and fine-tuning PM\textsubscript{10} models (which use ImageNet-derived features of RGB images) are about as accurate as the baseline and random models when using the six meteorological variables, and more accurate when using a high-dimensional representation of meteorology (Table \ref{tab:scores}). With a high-dimensional representation of meteorology, the transfer model captures almost half of the variation in PM\textsubscript{10} (R\textsuperscript{2} = 0.45, RMSE = 5.4 µg/m3, NMAE = 24\%). When using only image features, the transfer model is somewhat more accurate than the random model, with test set R\textsuperscript{2} of 0.23, RMSE of 6.4 µg/m3, and NMAE of 28\%. Using four-channel TOAR images slightly increases accuracy when using only image features (Table \ref{sup:tab:scores-toar}).

\begin{figure}
    \includegraphics[width=0.9\linewidth]{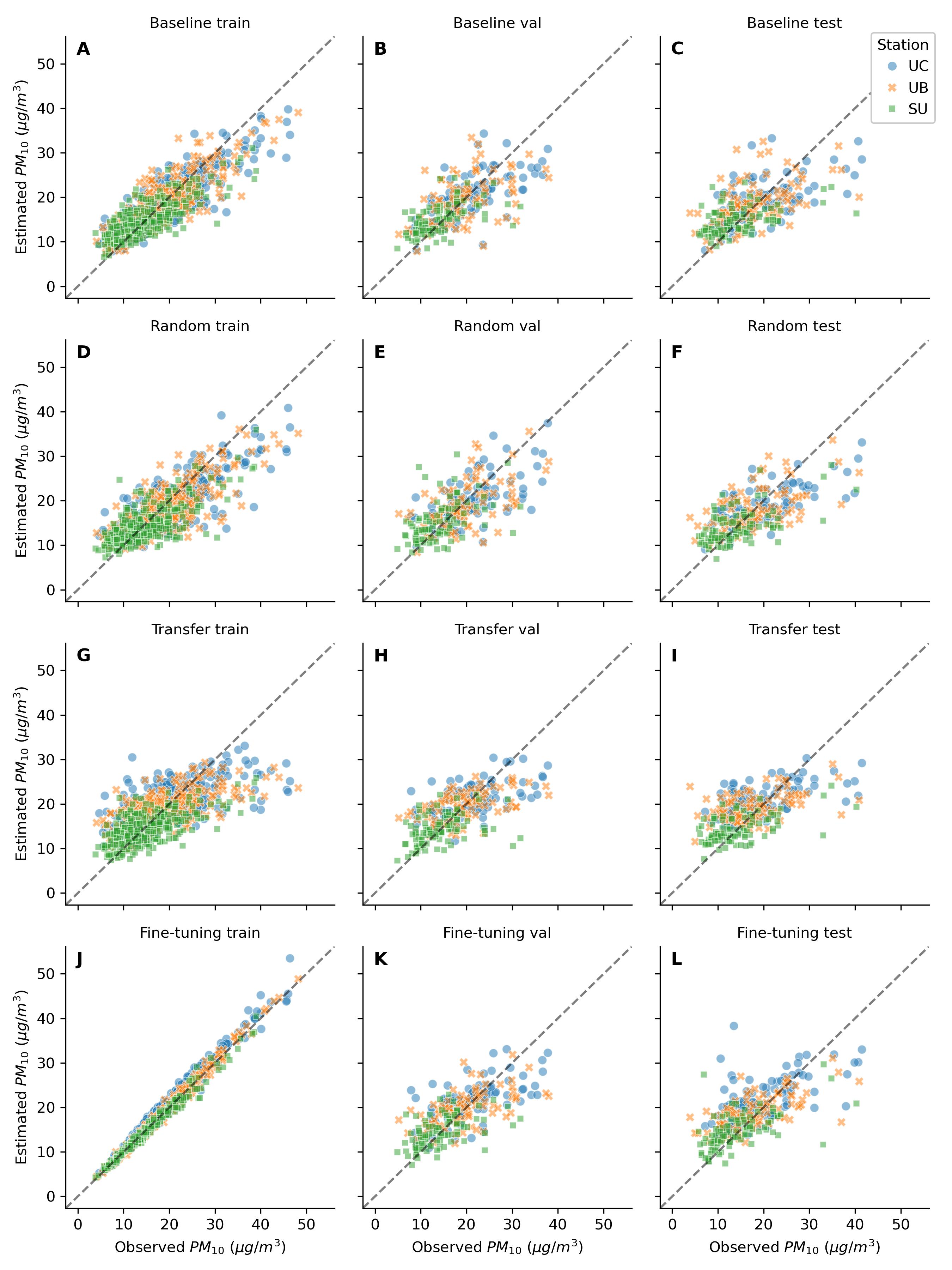}
    \caption{Observed and estimated PM\textsubscript{10} for the training (left column), validation (middle column), and test (right column) sets. Top row (A-C): baseline model (meteorology only); second row (D-F): random model (meteorology and random features of RGB images); third row (G-I): transfer model (meteorology and ImageNet-derived features of RGB images); bottom row (J-L): fine-tuning model (meteorology and fine-tuned ImageNet-derived features of RGB images).}
    \label{fig:scatter-pm-met}
\end{figure}

The transfer model performs surprisingly poorly on the training set, with R\textsuperscript{2} of just 0.38 when using only image features (Figure \ref{fig:scatter-pm-met}, Figure \ref{sup:fig:scatter-pm}, Table \ref{sup:tab:scores-pm}). This may be due to the short training period, since validation loss reaches a minimum after only 8 epochs. The fine-tuning model takes 38 epochs to reach a validation loss minimum and is much more accurate on the training set (R\textsuperscript{2} = 0.97). But the fine-tuning model performs similarly to the transfer model on the test set (Table \ref{tab:scores}). Taken together, these results indicate that image features are only weakly related to PM\textsubscript{10} mass concentration but do contain information that is complimentary to meteorology. As with the OP\textsubscript{DTT} models, using a similar dimension to represent the images and meteorology improves the model’s ability to leverage the multi-modal data.

\subsubsection{PM\textsubscript{10}: contrastive learning}

As with OP\textsubscript{AA} and OP\textsubscript{DTT}, the SimSiam PM\textsubscript{10} model (which uses contrastive features of RGB images selected by SimSiam)is less accurate than the models that use ImageNet-derived features (Table \ref{tab:scores}). When using only image features, the SimSiam model is slightly less accurate than the random model, with test set R\textsuperscript{2} of 0.12, RMSE of 6.8 µg/m3, and NMAE of 31\%. Although using four-channel TOAR images improves accuracy (R\textsuperscript{2} = 0.19, RMSE = 6.6 µg/m3, NMAE = 30\%) (Table \ref{sup:tab:scores-toar}), it fails to match the performance of the transfer and fine-tuning models. The SimSiam DL model does not match the performance of the transfer and fine-tuning models, and adapting contrastive features learned in Beijing results in very poor performance (Table \ref{tab:scores}).

Supplementary Figures \ref{sup:fig:loss-pm-met} and \ref{sup:fig:loss-pm-nomet} show training and validation loss curves for the PM\textsubscript{10} models.

\subsection{Comparison with previous work}

To the best of our knowledge, this is the first study to evaluate the capacity of satellite images to directly estimate 24-hour OP of PM\textsubscript{10} at a 1~km spatial resolution. A few recent studies have reported good performance estimating 24-hour OP from PM sources or components. \citet{Borlaza2021} estimated 24-hour OP of PM\textsubscript{10} in Grenoble based on the mass contribution of PM\textsubscript{10} sources derived from receptor models, achieving R\textsuperscript{2} of 0.77 for OP\textsubscript{AA} and 0.67 for OP\textsubscript{DTT}. \citet{Daellenbach2020} estimated 24-hour OP of PM\textsubscript{10} and PM\textsubscript{2.5} across Europe at approximately 20~km spatial resolution based on the mass contributions of PM\textsubscript{10} and PM\textsubscript{2.5} sources derived from a chemical transport model. The OP model was calibrated against 109 24-hour OP measures at five locations and achieved R\textsuperscript{2} of 0.71 for OP\textsubscript{AA} and 0.83 for OP\textsubscript{DTT} on 114 24-hour OP measures at two test locations. \citet{Xu2021} estimated annual mean OP of PM\textsubscript{2.5} across Canada at approximately 1~km spatial resolution based on the mass concentration of 14 PM\textsubscript{2.5} components, which were themselves estimated from satellite-derived AOD and a chemical transport model. The OP model was calibrated using annual mean OP at 33 locations and achieved leave-one-location-out R\textsuperscript{2} of 0.63 and RMSE of 8.8~pmol~min\textsuperscript{-1}m\textsuperscript{-3} (about 20\% of the median observation) for OP\textsubscript{AA}. For OP\textsubscript{DTT}, the model achieved R\textsuperscript{2} of 0.74 and RMSE of 8.77~pmol~min\textsuperscript{-1}~m\textsuperscript{-3} (about 17\% of the median observation). While our OP estimates are less accurate than these studies, our model captures about 50\% of the variability in OP\textsubscript{AA} and about 35\% of the variability in OP\textsubscript{DTT} using only meteorology or satellite images. If these results are confirmed in other areas, then it may be possible to roughly estimate OP from readily available data without needing detailed PM speciation.

A few previous studies used land use regression to estimate spatial variation in weekly \citep{Yanosky2012}, seasonal \citep{Weichenthal2019}, or annual mean OP \citep{Gulliver2018, Hellack2017, Jedynska2017, Yang2015} across spatial extents ranging from a city to a large country. Accuracy varied widely, with no model possible in some areas and leave-one-location-out R\textsuperscript{2} ranging from 0.32 to 0.82 for OP\textsubscript{AA} and 0.05 to 0.59 for OP\textsubscript{DTT} in areas where modelling succeeded. Most studies reported that OP was more spatially variable than PM mass concentration. Traffic-related variables were consistently the main predictors of OP, and most studies reported a lack of good spatial predictors other than fuel combustion and vehicle brake and tire wear. We used OP measures from three closely located and moderately correlated stations, so our main focus was on temporal variation in OP rather than spatial variation. If we compare the average estimate with the average measure at each station, our fine-tuning model using only satellite images achieves high accuracy (NMAE of 9\% for OP\textsubscript{AA} and 4\% for each of OP\textsubscript{DTT} and PM\textsubscript{10}). This suggests that it may be possible to accurately estimate typical spatial patterns of OP from satellite images. \citet{Yanosky2012} also reported higher accuracy when estimating annual rather than weekly mean OP in London via land use regression, and \citet{Zheng2020} and \citet{Jiang2022} reported higher accuracy estimating annual rather than 24-hour PM\textsubscript{2.5} in Delhi and Beijing using an approach and data similar to ours.

Our PM\textsubscript{10} models perform less well (R\textsuperscript{2} = 0.37, NMAE = 25\% for the fine-tuning model that uses both meteorology and image features) than a previous study in France that estimated 24-hour PM\textsubscript{10} mass concentration from satellite-derived AOD, achieving R\textsuperscript{2} of 0.77 and NMAE of 19\% for the region of France containing Grenoble \citep{Hough2021}. This may be because our models use fewer predictive variables and less training data, with only five years of daily PM\textsubscript{10} measures at three locations compared to the 20 years and 330 locations used by \citet{Hough2021}. Previous studies that used the same type of satellite images and similar meteorological data as our study reported R\textsuperscript{2} of 0.81 and NMAE of 25\% when estimating 24-hour PM\textsubscript{2.5} at 35 locations in Beijing \citep{Zheng2020} and NMAE of 19\% (R\textsuperscript{2} not reported) when estimating 24-hour PM\textsubscript{2.5} at 51 locations in Delhi \citep{Zheng2021}. An alternate model using only satellite images was less accurate (R\textsuperscript{2} = 0.35, NMAE = 50\% in Beijing; R\textsuperscript{2} = 0.26, NMAE = 46\% in Delhi) \citep{Jiang2022}. Our fine-tuning PM\textsubscript{10} model that uses both meteorology and image features has similar relative errors (NMAE = 25\%) but captures less of the variation in PM\textsubscript{10} (R\textsuperscript{2} = 0.37) than these studies. This might be due to the cleaner and less variable air quality of our study area compared to Beijing and Delhi: small differences in PM\textsubscript{10} may be less visible in satellite images and less driven by meteorology.

\subsection{Limitations and recommendations}

Our study's main limitation is the size of our dataset. Since OP is not routinely monitored, our dataset of 1354 24-hour OP measures from three locations over five years represents one of the longest available OP timeseries. In contrast, studies of PM concentration often use thousands of measures from tens of locations per urban area. Since only 27\% of the OP measures had a corresponding cloud-free satellite image, we trained our OP models on just 210 satellite image-OP measurement pairs, reserving an additional 80 image-OP pairs for validation and 75 for testing. The small test set means that some differences in R\textsuperscript{2}, RMSE, and NMAE may be due to chance. For example, the test set estimates of the baseline OP\textsubscript{DTT} model are less accurate (RMSE = 0.78~nmol~min\textsuperscript{-1}~m\textsuperscript{-3}) than the estimates of the transfer model that uses both meteorology and image features (RMSE = 0.73~nmol~min\textsuperscript{-1}~m\textsuperscript{-3}). But the 95\% confidence interval of these scores, estimated across 1000 bootstrap samples of the test set estimates, overlaps (0.64 – 0.83~nmol~min\textsuperscript{-1}~m\textsuperscript{-3} for the baseline model; 0.56 – 0.79~nmol~min\textsuperscript{-1}~m\textsuperscript{-3} for the transfer model; Table \ref{sup:tab:boots}). The small number of training samples and locations may also have limited our model's ability to learn relationships between images and OP. Future studies should evaluate whether increasing the amount of training data improves performance. Future studies should also use OP measures from other areas and more locations to confirm the approach is broadly applicable and evaluate how well it captures spatial variation in OP.

Our model may also have been limited by the 30~km spatial resolution of the ERA5 boundary layer data, which cannot represent the $\sim$10~km wide valley in which Grenoble is located. We chose to use the ERA5 data despite its low spatial resolution because its global coverage could facilitate applying our model in other areas. More accurate boundary layer height data might benefit the models that use meteorological data, but would not affect the models that use only image features. Our OP\textsubscript{DTT} models might also have benefited from indicators of photochemical aging such as solar radiation or ozone concentration, since OP\textsubscript{DTT} is sensitive to the oxidation of secondary organic aerosols \citep{Bates2019, Gao2020}. Future work could explore whether additional variables improve the model estimates.

\section{Conclusions}

This study is the first to directly estimate the OP of PM from satellite data. Using a combination of meteorological variables and features of satellite images extracted by a state-of-the-art CNN, our models estimate 24-hour mean PM\textsubscript{10} mass concentration, OP\textsubscript{AA}, and OP\textsubscript{DTT} with mean absolute error that is 22-32\% of the mean observation. Image features alone capture 23\%, 49\%, and 36\% of the variation in PM\textsubscript{10}, OP\textsubscript{AA}, and OP\textsubscript{DTT}, respectively. Generic image features learned from the large ImageNet dataset are better predictors of PM\textsubscript{10} mass concentration and OP than contrastive features learned from satellite images. Our findings suggest that it may be possible to roughly estimate OP from readily available meteorological variables and satellite images using off-the-shelf CNN features with little or no tuning. If confirmed in other areas, this method could represent a relatively low-cost approach to expand the temporal or spatial coverage of OP estimates.

\subsection*{Code and data availability}

The code for this study is available at \url{https://gitlab.com/ihough/deep-satellite-op}. The trained model weights are available from \url{https://doi.org/10.57745/ZOYJHD}. The PM\textsubscript{10} mass concentration data are available from Atmo Auvergne-Rhône-Alpes at \url{https://api.atmo-aura.fr}. The OP\textsubscript{AA} and OP\textsubscript{DTT} data are available from the authors upon reasonable request. The meteorological data are openly available from the Copernicus Climate Data Store at \url{https://doi.org/10.24381/cds.e2161bac} (ERA5-Land) and \url{https://doi.org/10.24381/cds.adbb2d47} (ERA5). The satellite data are available from Planet Labs Inc. but restrictions apply; the data were used under license for the current study and so are not publicly available.

\subsection*{CRediT author statement}

\textbf{Ian Hough:} Conceptualization, Methodology, Software, Formal analysis, Data curation, Writing – original draft, Writing - review and editing, Visualization. \textbf{Loïc Argentier:} Software, Writing - review and editing. \textbf{Ziyang Jiang:} Software, Writing – review and editing. \textbf{Tongshu Zheng:} Writing – review and editing. \textbf{Mike Bergin:} Writing – review and editing. \textbf{David Carlson:} Writing – review and editing. \textbf{Jean-Luc Jaffrezo:} Conceptualization, Writing - review and editing. \textbf{Jocelyn Chanussot:} Conceptualization, Methodology, Resources, Writing - review and editing, Supervision, Funding acquisition. \textbf{Gaëlle Uzu:} Conceptualization, Methodology, Resources, Writing - review and editing, Supervision, Funding acquisition.

\subsection*{Declaration of competing interest}

The authors declare that they have no known competing financial interests or personal relationships that could have appeared to influence the work reported in this paper.

\subsection*{Acknowledgements}

The authors thank the many people at Atmo Auvergne-Rhône-Alpes and the IGE who contributed to collecting and analysing the samples.

\subsection*{Funding}

This work was supported by MIAI@Grenoble Alpes (ANR-19-P3IA-0003). PM sampling and chemical analyses were supported by the French National Research Agency’s Investissements d’avenir programme through the IDEX UGA (ANR-15-IDEX-0002), ACME, MobilAir, and Get OP Stand OP (ANR-19-CE34-0002) projects. Sampling and analyses were also supported by the Fondation Université Grenoble Alpes through the Predict’Air project, by ADEME (convention 1662C0029) through the QAMECS project, and by LCSQA and the French Ministry of Environment through the CARA programme. Chemical analyses were performed at the IGE Air-O-Sol facility whose equipment was funded in part by the Labex OSUG@2020 (ANR-10-LABX-0056) project.

\bibliographystyle{elsarticle-harv} 
\bibliography{references}

\clearpage

\makeatletter
\renewcommand\appendix{
  \setcounter{section}{0}%
  \setcounter{subsection}{0}%
  \setcounter{table}{0}%
  \setcounter{figure}{0}%
  \gdef\thefigure{S\arabic{figure}}%
  \gdef\thetable{S\arabic{table}}%
  \gdef\thesection{\appendixname\@Alph\c@section}%
}
\makeatother

\appendix
\section{Filter used to exclude cloudy PlanetScope images}
\label{sup:filter}

\begin{lstlisting}[language=SQL, basicstyle=\ttfamily\scriptsize]
(
    cover == 1 AND cloud_cover == 0
    AND green_q05 < 0.25
    AND NOT date in ['2018-08-18', '2021-01-07', '2021-06-25']
    AND NOT (date in ['2019-12-06', '2020-01-08'] AND station != 'SU')
    AND NOT (date == '2020-05-10' AND instrument == 'PS2')
    AND NOT (
        date == '2021-08-12' AND instrument == 'PS2' AND station == 'UC'
    )
) OR (
    cover == 1 AND cloud_cover > 0 AND cloud_cover < 1
    AND green_q05 < 0.25 AND green_q50 < 0.26 AND green_q95 < 0.5
    AND date in [
        '2017-01-19', '2017-01-26', '2017-03-15', '2017-03-17',
        '2017-04-05', '2017-07-08', '2017-07-28', '2017-08-09',
        '2017-10-22', '2018-02-13', '2018-03-08', '2018-03-16',
        '2018-03-19', '2018-03-23', '2018-03-24', '2018-03-26',
        '2018-04-25', '2018-04-26', '2018-04-27', '2018-04-28',
        '2018-05-05', '2018-05-07', '2018-05-11', '2018-05-12',
        '2018-05-21', '2018-06-01', '2018-06-02', '2018-06-08',
        '2018-06-09', '2018-06-24', '2018-08-06', '2018-08-19',
        '2018-09-22', '2018-09-30', '2018-10-14', '2019-03-16',
        '2019-04-01', '2019-04-20', '2019-04-22', '2019-04-23',
        '2019-04-29', '2019-07-21', '2019-08-05', '2019-08-19',
        '2019-09-11', '2019-09-27', '2019-10-14', '2019-10-24',
        '2020-01-13', '2020-02-25', '2020-06-15', '2020-06-21',
        '2020-07-02', '2020-09-02', '2020-10-03', '2021-03-25',
        '2021-06-10', '2021-07-07', '2021-07-24', '2021-08-28',
        '2021-09-22', '2021-09-29', '2021-10-26'
    ]
) OR (cover > 0.7 AND cover < 1 AND green_q95 < 0.21)}
\end{lstlisting}

Variable definitions:

\begin{itemize}
    \item \texttt{cover}: fraction of 1~km\textsuperscript{2} covered by image
    \item \texttt{green\_q05}: 5\textsuperscript{th} percentile of green band reflectance across all images
    \item \texttt{green\_q50}: median green band reflectance across all images
    \item \texttt{green\_q95}: 95\textsuperscript{th} percentile of green band reflectance across all images
\end{itemize}

\clearpage

\section{Supplementary Figures}

\begin{figure}[!b]
    \includegraphics[width=1\linewidth]{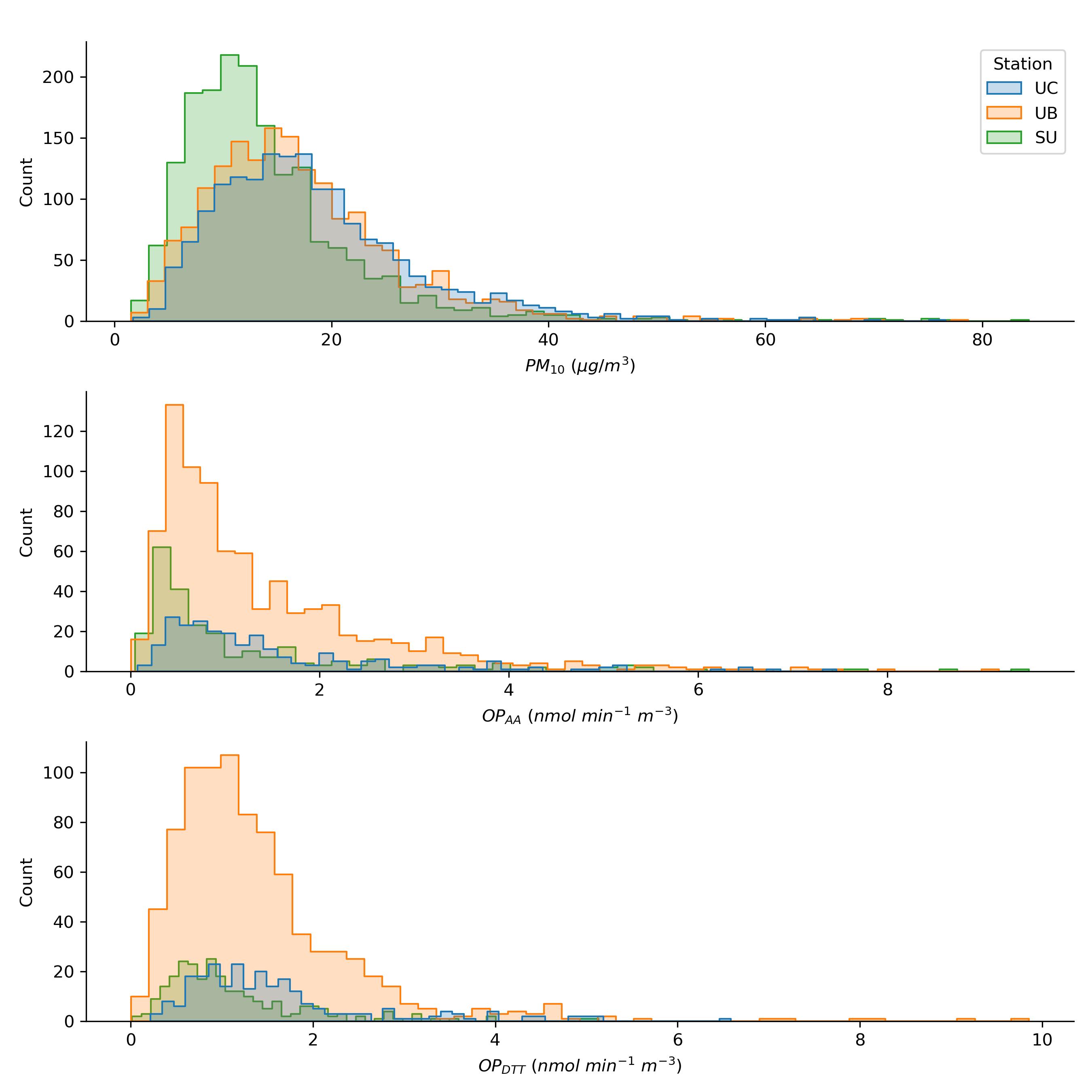}
    \caption{Distribution of 24-hour PM\textsubscript{10}, OP\textsubscript{AA}, and OP\textsubscript{DTT} observations. Vertical dashed line shows outlier cutoff (measures higher than this value were excluded from analyses).}
    \label{sup:fig:aq-dist}
\end{figure}

\begin{figure}
    \includegraphics[width=1\linewidth]{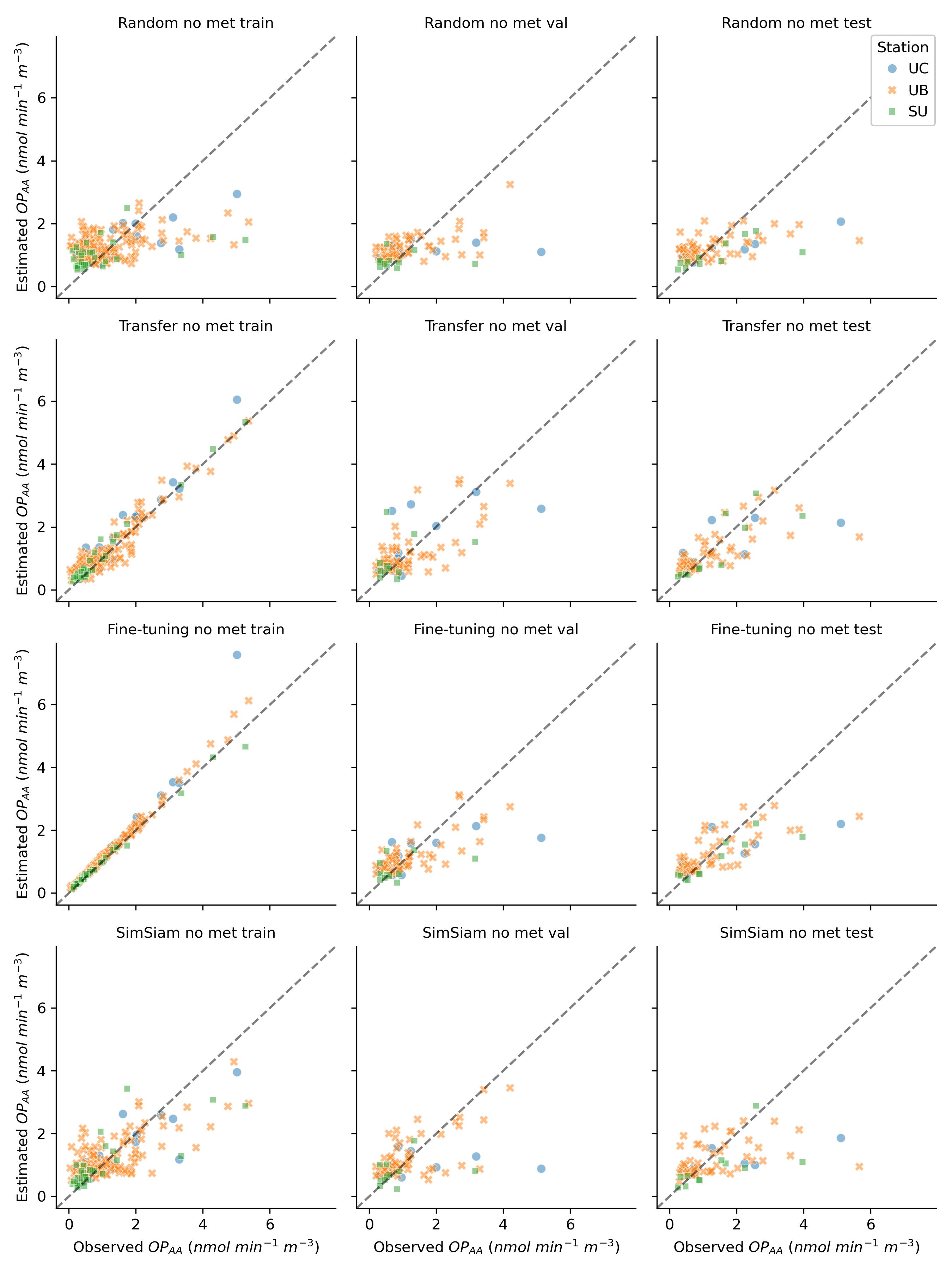}
    \caption{Observed and estimated OP\textsubscript{AA} from models that use only RGB image features}
    \label{sup:fig:scatter-aa}
\end{figure}

\begin{figure}
    \includegraphics[width=0.9\linewidth]{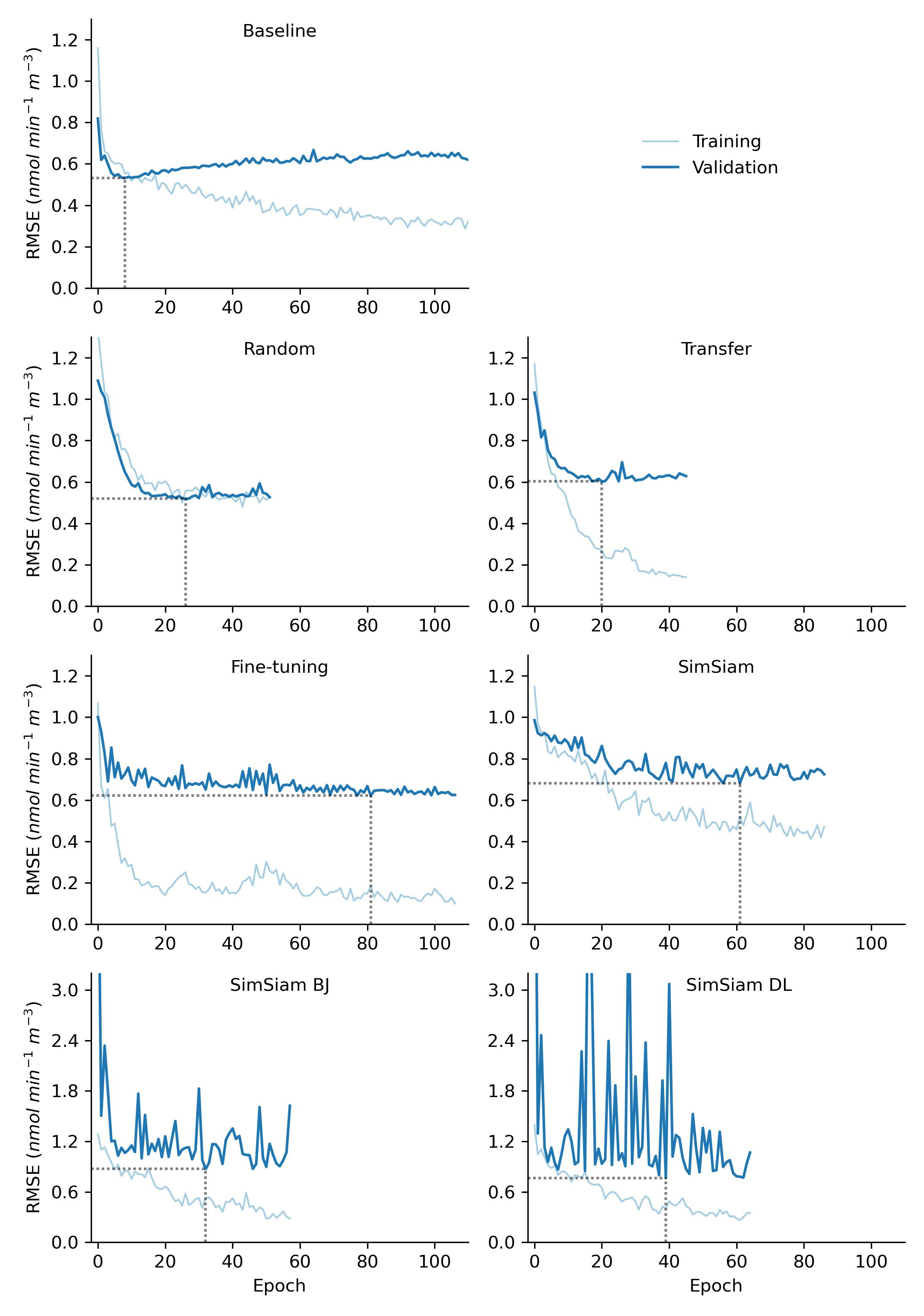}
    \caption{Training and validation loss of OP\textsubscript{AA} models that use meteorology. Dotted lines indicate the epoch with minimum validation loss.}
    \label{sup:fig:loss-aa-met}
\end{figure}

\begin{figure}
    \includegraphics[width=1\linewidth]{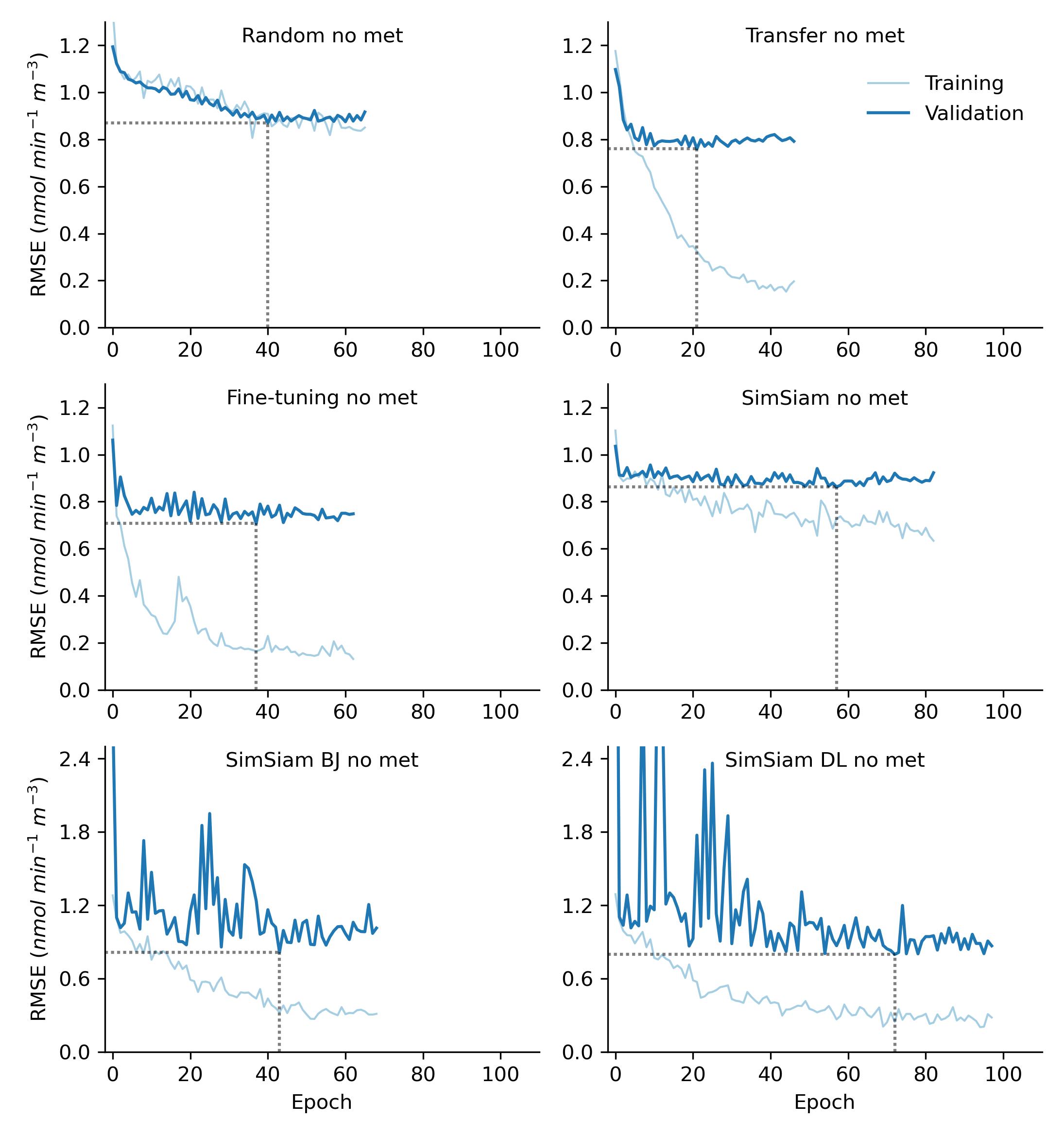}
    \caption{Training and validation loss of OP\textsubscript{AA} models that use meteorology. Dotted lines indicate the epoch with minimum validation loss.}
    \label{sup:fig:loss-aa-nomet}
\end{figure}

\begin{figure}
    \includegraphics[width=1\linewidth]{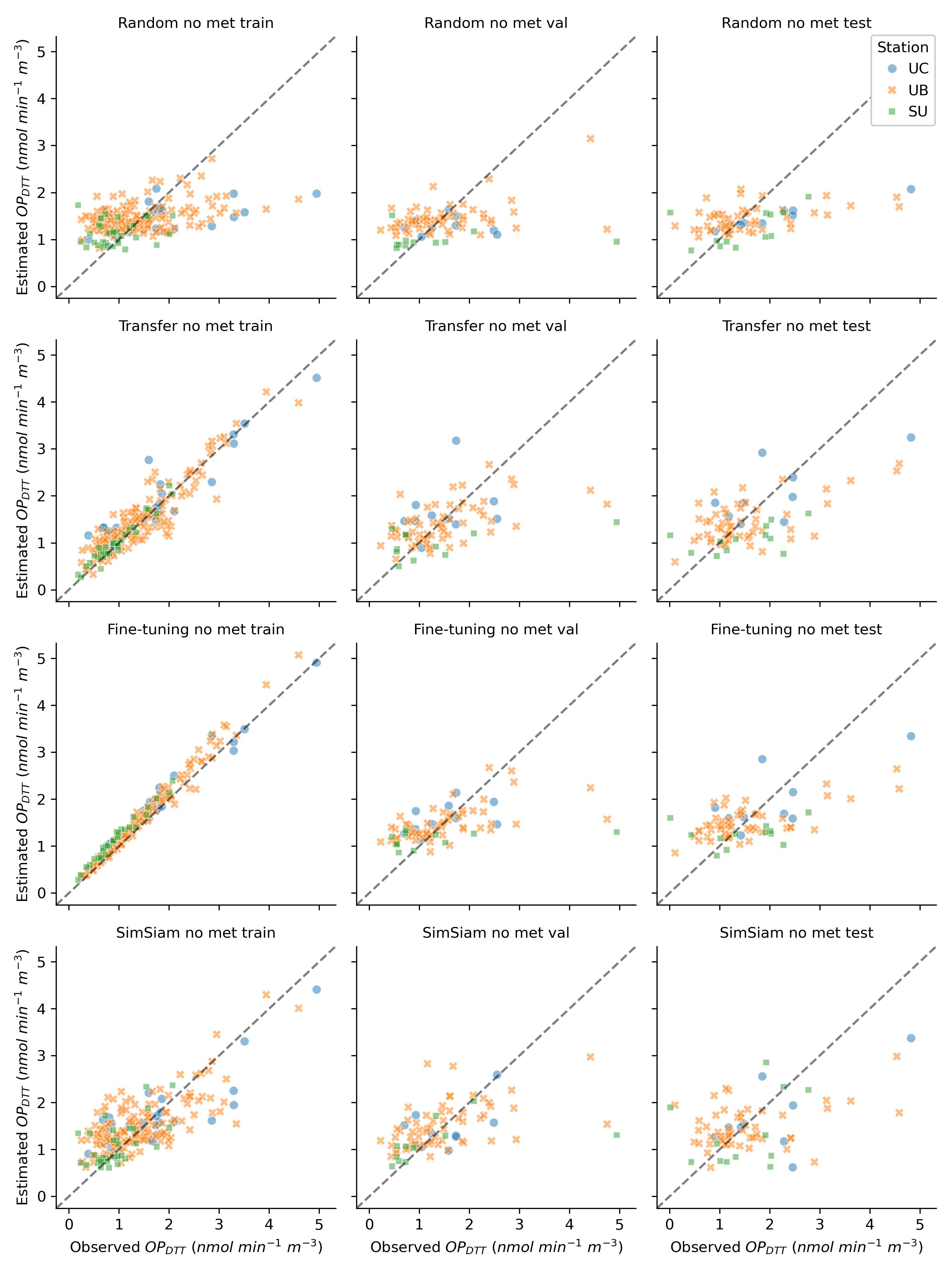}
    \caption{Observed and estimated OP\textsubscript{DTT} from models that use only RGB image features}
    \label{sup:fig:scatter-dtt}
\end{figure}

\begin{figure}
    \includegraphics[width=0.9\linewidth]{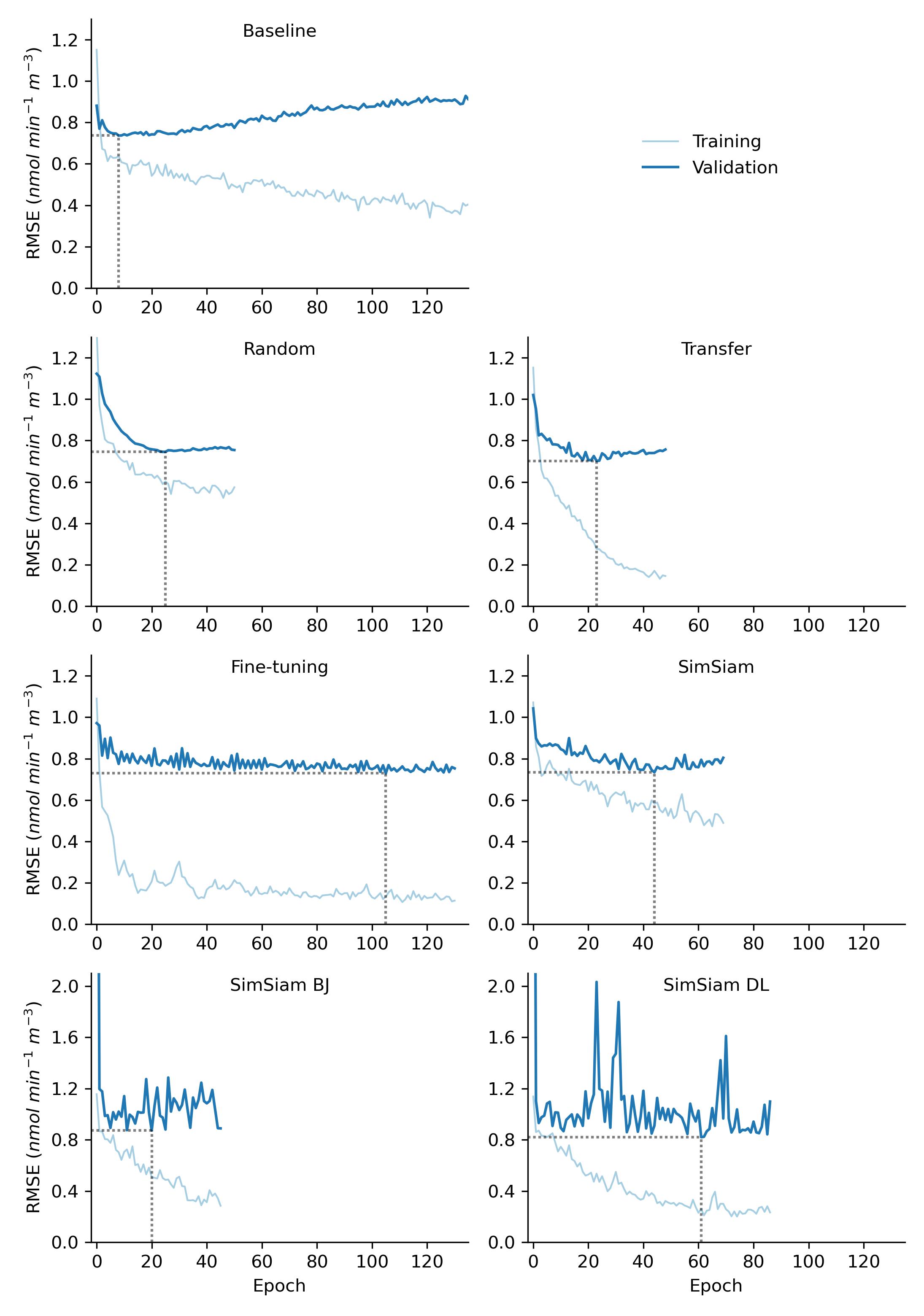}
    \caption{Training and validation loss of OP\textsubscript{DTT} models that use meteorology. Dotted lines indicate the epoch with minimum validation loss.}
    \label{sup:fig:loss-dtt-met}
\end{figure}

\begin{figure}
    \includegraphics[width=1\linewidth]{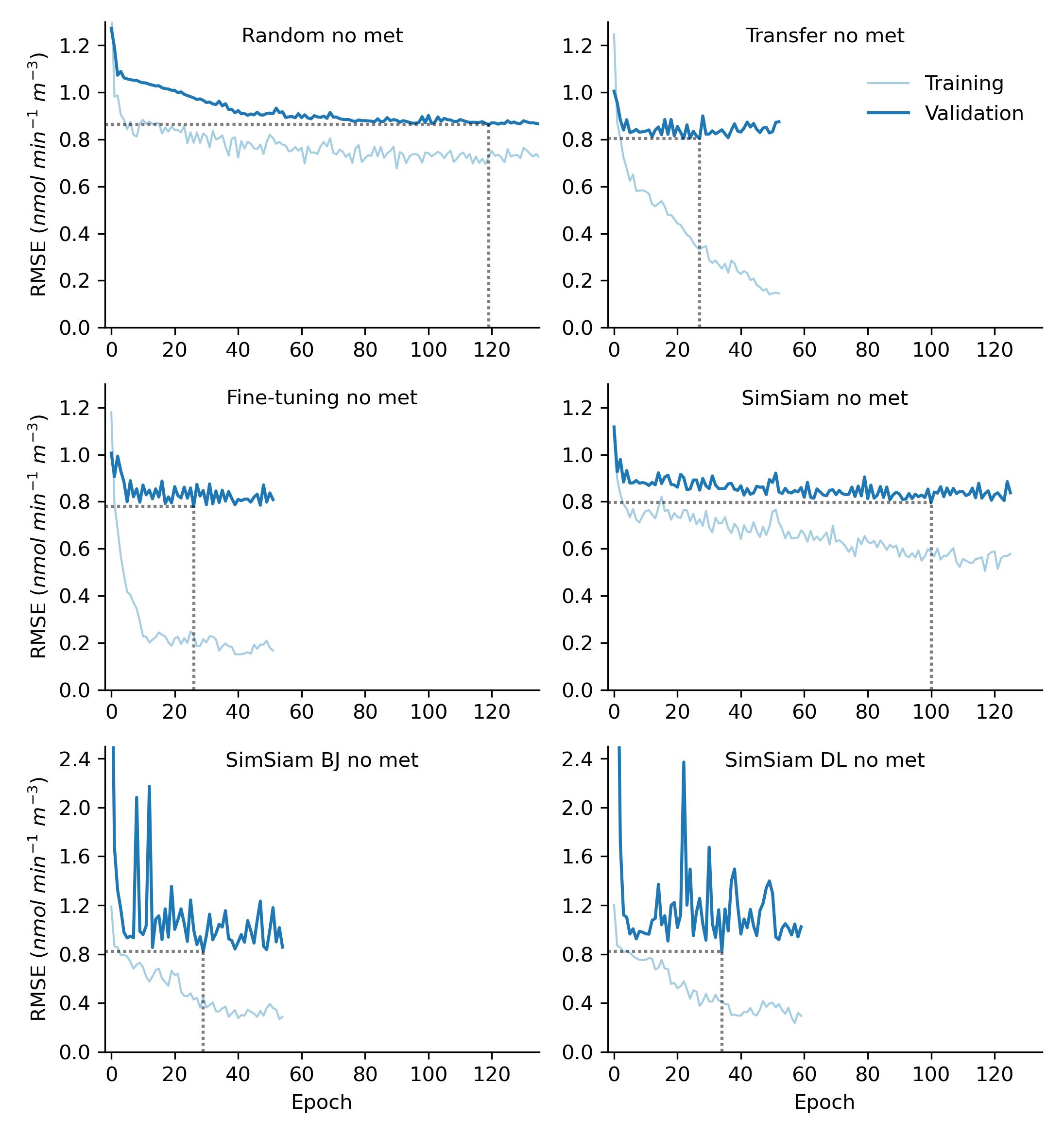}
    \caption{Training and validation loss of OP\textsubscript{DTT} models that use meteorology. Dotted lines indicate the epoch with minimum validation loss.}
    \label{sup:fig:loss-dtt-nomet}
\end{figure}

\begin{figure}
    \includegraphics[width=1\linewidth]{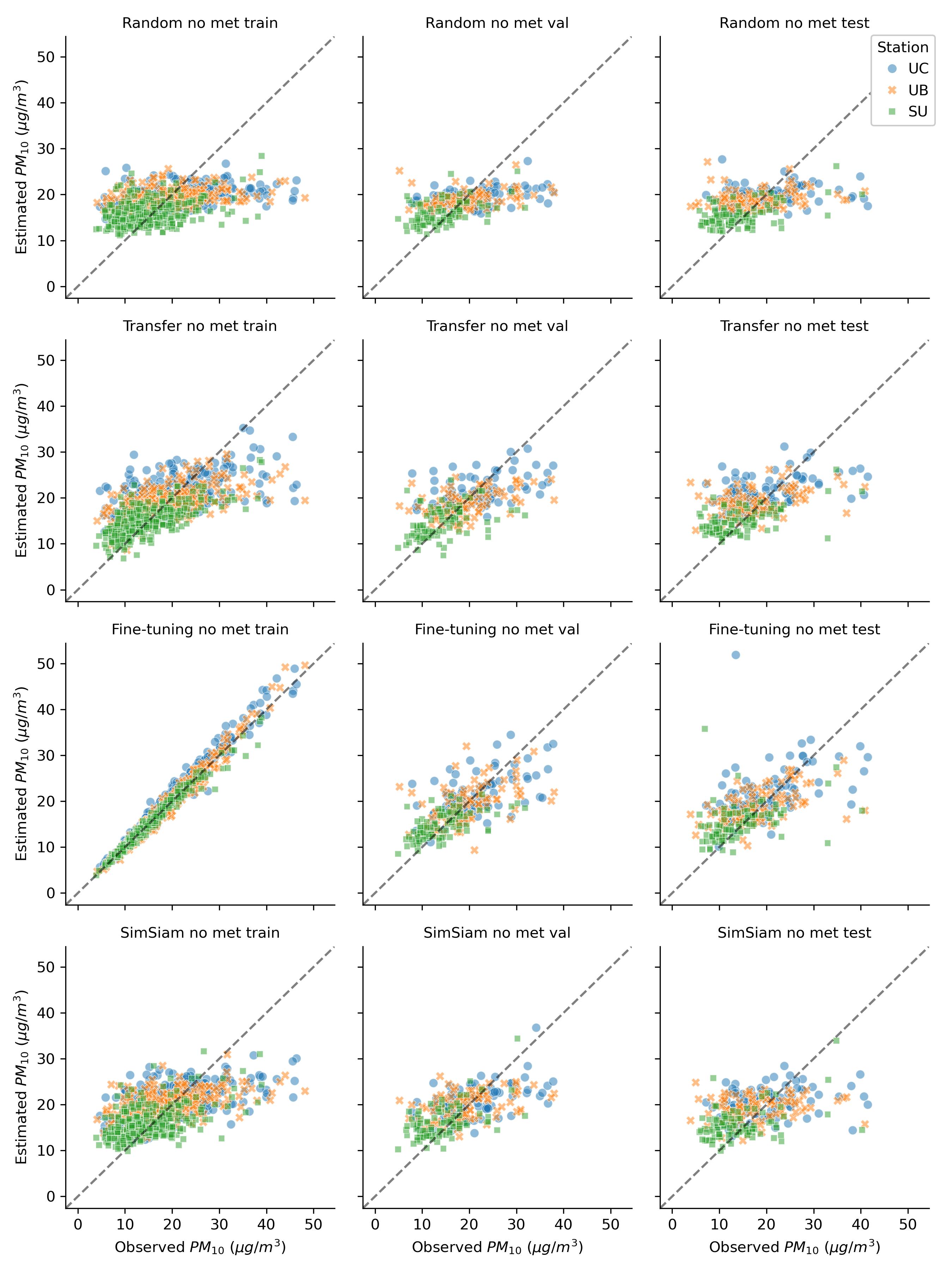}
    \caption{Observed and estimated PM\textsubscript{10} from models that use only RGB image features}
    \label{sup:fig:scatter-pm}
\end{figure}

\begin{figure}
    \includegraphics[width=0.9\linewidth]{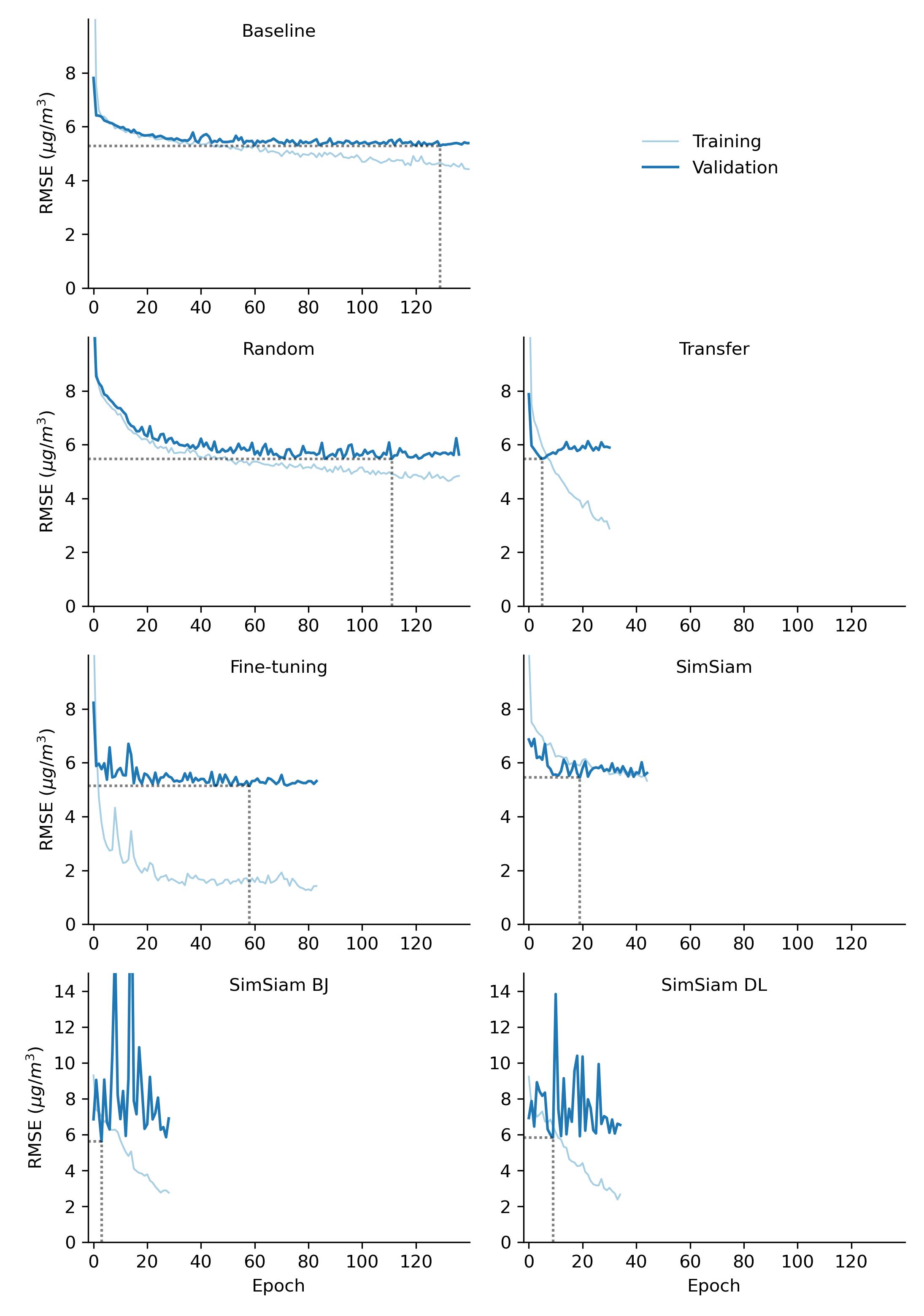}
    \caption{Training and validation loss of PM\textsubscript{10} models that use meteorology. Dotted lines indicate the epoch with minimum validation loss.}
    \label{sup:fig:loss-pm-met}
\end{figure}

\begin{figure}
    \includegraphics[width=1\linewidth]{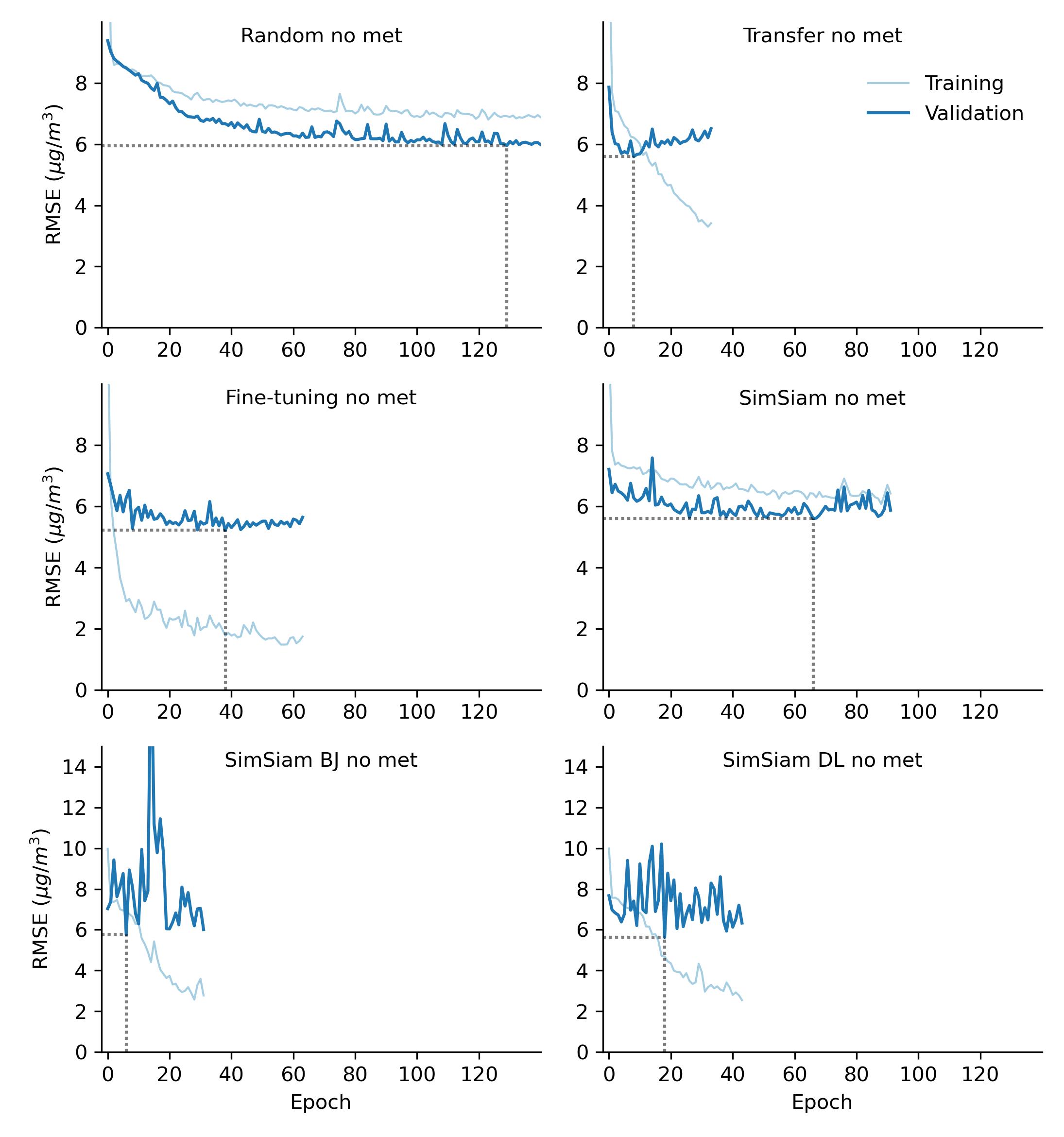}
    \caption{Training and validation loss of OP\textsubscript{PM} models that use meteorology. Dotted lines indicate the epoch with minimum validation loss.}
    \label{sup:fig:loss-pm-nomet}
\end{figure}

\clearpage

\section{Supplementary Tables}

\begin{table}[h]
    \begin{threeparttable}
        \caption{Spearman correlation of air quality measures between stations}
        \label{sup:tab:station-cor}
        \small
        \begin{tabular}{llrrrrrrrrr}
            \toprule
            & & \multicolumn{2}{c}{PM\textsubscript{10}} & & \multicolumn{2}{c}{OP\textsubscript{AA}} & & \multicolumn{2}{c}{OP\textsubscript{DTT}} \\
            \cmidrule(lr){3-4}
            \cmidrule(lr){6-7}
            \cmidrule(lr){9-10}
            Station & & UC & UB & & UC & UB & & UC & UB \\
            \midrule
            UB & & 0.94 & & & 0.85 & & & 0.67 & & \\
            SU & & 0.88 & 0.89 & & 0.81 & 0.86 & & 0.59 & 0.58 \\
            \bottomrule
        \end{tabular}
        \begin{tablenotes}
            \item UC = Urban center; BG = Urban background; SU = Suburban.
        \end{tablenotes}
    \end{threeparttable}
\end{table}

\begin{table}[h]
    \begin{threeparttable}
        \caption{Test set performance of the OP\textsubscript{AA}, OP\textsubscript{DTT}, and PM\textsubscript{10} models using TOAR images.}
        \label{sup:tab:scores-toar}
        \small
        \begin{tabular}{llrrrrrrrrr}
            \toprule
            \multicolumn{2}{c}{} & \multicolumn{3}{c}{OP\textsubscript{AA}} & \multicolumn{3}{c}{OP\textsubscript{DTT}} & \multicolumn{3}{c}{PM\textsubscript{10}} \\
            \cmidrule(lr){3-5}
            \cmidrule(lr){6-8}
            \cmidrule(lr){9-11}
            Model & Features & R\textsuperscript{2} & RMSE & NMAE & R\textsuperscript{2} & RMSE & NMAE & R\textsuperscript{2} & RMSE & NMAE \\
            \midrule
            Transfer    & I+M& 0.48 & 0.80 & 37\% & 0.32 & 0.78 & 37\% & 0.34 & 5.9 & 26\% \\
            Fine-tuning & I+M& 0.53 & 0.77 & 35\% & 0.42 & 0.71 & 35\% & 0.27 & 6.2 & 24\% \\
            SimSiam     & I+M& 0.36 & 0.89 & 44\% & 0.34 & 0.76 & 35\% & 0.32 & 6.0 & 27\% \\
            \addlinespace
            Transfer    & I  & 0.47 & 0.81 & 38\% & 0.32 & 0.77 & 38\% & 0.29 & 6.2 & 26\% \\
            Fine-tuning & I  & 0.53 & 0.77 & 37\% & 0.40 & 0.72 & 36\% & 0.25 & 6.3 & 26\% \\
            SimSiam     & I  & 0.35 & 0.90 & 44\% & 0.29 & 0.79 & 37\% & 0.19 & 6.6 & 30\% \\
            \bottomrule
        \end{tabular}
        \begin{tablenotes}
            \item I+M = TOAR image features (n = 2048) and meteorological variables (n = 6); I = TOAR image features. RMSE units are nmol~min\textsuperscript{-1}~m\textsuperscript{-3} for OP\textsubscript{AA} and OP\textsubscript{DTT} and µg/m\textsuperscript{3} for PM\textsubscript{10}.
        \end{tablenotes}
    \end{threeparttable}
\end{table}

\begin{table}
    \begin{threeparttable}
        \caption{Performance of the OP\textsubscript{AA} models.}
        \label{sup:tab:scores-aa}
        \small
        \begin{tabular}{llrrrrrrrrr}
            \toprule
            \multicolumn{2}{c}{} & \multicolumn{3}{c}{Train set} & \multicolumn{3}{c}{Validation set} & \multicolumn{3}{c}{Test set} \\
            \cmidrule(lr){3-5}
            \cmidrule(lr){6-8}
            \cmidrule(lr){9-11}
            Model & Features & R\textsuperscript{2} & RMSE & NMAE & R\textsuperscript{2} & RMSE & NMAE & R\textsuperscript{2} & RMSE & NMAE \\
            \midrule
            Baseline    & M  & 0.71 & 0.54 & 33\% & 0.71 & 0.53 & 31\% & 0.60 & 0.71 & 35\% \\
            Random      & I+M& 0.72 & 0.53 & 32\% & 0.72 & 0.52 & 28\% & 0.55 & 0.75 & 36\% \\
            \addlinespace
            Transfer    & I+M& 0.95 & 0.21 & 15\% & 0.63 & 0.60 & 38\% & 0.54 & 0.76 & 33\% \\
            Fine-tuning & I+M& 0.99 & 0.11 &  5\% & 0.60 & 0.62 & 35\% & 0.60 & 0.71 & 34\% \\
            Transfer    & I+H& 0.89 & 0.33 & 21\% & 0.71 & 0.54 & 30\% & 0.62 & 0.69 & 31\% \\
            Fine-tuning & I+H& 0.98 & 0.14 &  9\% & 0.71 & 0.53 & 30\% & 0.63 & 0.68 & 31\% \\
            \addlinespace
            SimSiam     & I+M& 0.82 & 0.43 & 29\% & 0.52 & 0.68 & 39\% & 0.44 & 0.84 & 40\% \\
            SimSiam BJ  & I+M& 0.67 & 0.57 & 38\% & 0.22 & 0.87 & 48\% & 0.15 & 1.03 & 50\% \\
            SimSiam DL  & I+M& 0.89 & 0.33 & 21\% & 0.40 & 0.77 & 41\% & 0.42 & 0.86 & 43\% \\
            \addlinespace
            Random      & I  & 0.21 & 0.89 & 57\% & 0.22 & 0.87 & 50\% & 0.23 & 0.98 & 48\% \\
            Transfer    & I  & 0.90 & 0.31 & 21\% & 0.40 & 0.76 & 46\% & 0.48 & 0.81 & 36\% \\
            Fine-tuning & I  & 0.95 & 0.23 & 11\% & 0.48 & 0.71 & 40\% & 0.49 & 0.80 & 38\% \\
            SimSiam     & I  & 0.56 & 0.67 & 41\% & 0.23 & 0.86 & 46\% & 0.14 & 1.04 & 48\% \\
            SimSiam BJ  & I  & 0.89 & 0.33 & 21\% & 0.31 & 0.82 & 47\% & 0.17 & 1.02 & 50\% \\
            SimSiam DL  & I  & 0.96 & 0.20 & 14\% & 0.34 & 0.80 & 43\% & 0.45 & 0.83 & 41\% \\
            \bottomrule
        \end{tabular}
        \begin{tablenotes}
            \item M = meteorological variables (n = 6); I+M = RGB image features (n = 2048) and meteorological variables; I = RGB image features. RMSE units are nmol~min\textsuperscript{-1}~m\textsuperscript{-3}.
        \end{tablenotes}
    \end{threeparttable}
\end{table}

\begin{table}
    \begin{threeparttable}
        \caption{Performance of the OP\textsubscript{DTT} models.}
        \label{sup:tab:scores-dtt}
        \small
        \begin{tabular}{llrrrrrrrrr}
            \toprule
            \multicolumn{2}{c}{} & \multicolumn{3}{c}{Train set} & \multicolumn{3}{c}{Validation set} & \multicolumn{3}{c}{Test set} \\
            \cmidrule(lr){3-5}
            \cmidrule(lr){6-8}
            \cmidrule(lr){9-11}
            Model & Features & R\textsuperscript{2} & RMSE & NMAE & R\textsuperscript{2} & RMSE & NMAE & R\textsuperscript{2} & RMSE & NMAE \\
            \midrule
            Baseline    & M  & 0.45 & 0.59 & 30\% & 0.34 & 0.74 & 37\% & 0.30 & 0.78 & 37\% \\
            Random      & I+M& 0.46 & 0.59 & 30\% & 0.32 & 0.75 & 35\% & 0.29 & 0.79 & 37\% \\
            \addlinespace
            Transfer    & I+M& 0.89 & 0.26 & 15\% & 0.40 & 0.70 & 33\% & 0.39 & 0.73 & 36\% \\
            Fine-tuning & I+M& 0.97 & 0.13 &  6\% & 0.35 & 0.73 & 32\% & 0.40 & 0.73 & 35\% \\
            Transfer    & I+H& 0.70 & 0.44 & 24\% & 0.36 & 0.72 & 35\% & 0.48 & 0.67 & 32\% \\
            Fine-tuning & I+H& 0.96 & 0.15 &  8\% & 0.39 & 0.71 & 33\% & 0.48 & 0.67 & 33\% \\
            \addlinespace
            SimSiam     & I+M& 0.52 & 0.55 & 31\% & 0.34 & 0.73 & 35\% & 0.24 & 0.82 & 40\% \\
            SimSiam BJ  & I+M& 0.44 & 0.59 & 35\% & 0.07 & 0.87 & 40\% & 0.10 & 0.89 & 42\% \\
            SimSiam DL  & I+M& 0.89 & 0.27 & 15\% & 0.18 & 0.82 & 37\% & 0.31 & 0.78 & 38\% \\
            \addlinespace
            Random      & I  & 0.20 & 0.71 & 39\% & 0.09 & 0.86 & 40\% & 0.18 & 0.85 & 38\% \\
            Transfer    & I  & 0.85 & 0.31 & 17\% & 0.21 & 0.80 & 38\% & 0.35 & 0.76 & 37\% \\
            Fine-tuning & I  & 0.94 & 0.19 & 11\% & 0.26 & 0.78 & 35\% & 0.36 & 0.75 & 37\% \\
            SimSiam     & I  & 0.54 & 0.54 & 31\% & 0.23 & 0.80 & 36\% & 0.15 & 0.87 & 40\% \\
            SimSiam BJ  & I  & 0.80 & 0.36 & 19\% & 0.17 & 0.82 & 38\% &-0.01 & 0.94 & 45\% \\
            SimSiam DL  & I  & 0.76 & 0.39 & 22\% & 0.17 & 0.82 & 38\% & 0.23 & 0.82 & 40\% \\
            \bottomrule
        \end{tabular}
        \begin{tablenotes}
            \item M = meteorological variables (n = 6); I+M = RGB image features (n = 2048) and meteorological variables; I = RGB image features. RMSE units are nmol~min\textsuperscript{-1}~m\textsuperscript{-3}.
        \end{tablenotes}
    \end{threeparttable}
\end{table}

\begin{table}
    \begin{threeparttable}
        \caption{Performance of the PM\textsubscript{10} models.}
        \label{sup:tab:scores-pm}
        \small
        \begin{tabular}{llrrrrrrrrr}
            \toprule
            \multicolumn{2}{c}{} & \multicolumn{3}{c}{Train set} & \multicolumn{3}{c}{Validation set} & \multicolumn{3}{c}{Test set} \\
            \cmidrule(lr){3-5}
            \cmidrule(lr){6-8}
            \cmidrule(lr){9-11}
            Model & Features & R\textsuperscript{2} & RMSE & NMAE & R\textsuperscript{2} & RMSE & NMAE & R\textsuperscript{2} & RMSE & NMAE \\
            \midrule
            Baseline    & M  & 0.69 & 4.2 & 18\% & 0.42 & 5.3 & 22\% & 0.33 & 5.9 & 27\% \\
            Random      & I+M& 0.60 & 4.7 & 20\% & 0.37 & 5.5 & 23\% & 0.43 & 5.5 & 25\% \\
            \addlinespace
            Transfer    & I+M& 0.42 & 5.7 & 24\% & 0.37 & 5.5 & 22\% & 0.31 & 6.0 & 27\% \\
            Fine-tuning & I+M& 0.98 & 1.1 &  5\% & 0.45 & 5.1 & 22\% & 0.37 & 5.8 & 25\% \\
            Transfer    & I+H& 0.90 & 2.4 & 10\% & 0.46 & 5.1 & 22\% & 0.45 & 5.4 & 24\% \\
            Fine-tuning & I+H& 0.98 & 1.0 &  4\% & 0.50 & 4.9 & 20\% & 0.44 & 5.4 & 22\% \\
            \addlinespace
            SimSiam     & I+M& 0.45 & 5.6 & 24\% & 0.38 & 5.5 & 23\% & 0.29 & 6.2 & 28\% \\
            SimSiam BJ  & I+M& 0.20 & 6.7 & 29\% & 0.34 & 5.6 & 24\% & 0.12 & 6.8 & 30\% \\
            SimSiam DL  & I+M& 0.34 & 6.1 & 26\% & 0.29 & 5.8 & 25\% & 0.16 & 6.7 & 30\% \\
            \addlinespace
            Random      & I  & 0.20 & 6.7 & 29\% & 0.26 & 6.0 & 25\% & 0.16 & 6.7 & 30\% \\
            Transfer    & I  & 0.38 & 5.9 & 25\% & 0.35 & 5.6 & 23\% & 0.23 & 6.4 & 28\% \\
            Fine-tuning & I  & 0.97 & 1.3 &  5\% & 0.43 & 5.2 & 21\% & 0.22 & 6.4 & 26\% \\
            SimSiam     & I  & 0.33 & 6.2 & 26\% & 0.34 & 5.6 & 24\% & 0.12 & 6.8 & 31\% \\
            SimSiam BJ  & I  & 0.16 & 6.9 & 31\% & 0.31 & 5.8 & 25\% & 0.09 & 6.9 & 32\% \\
            SimSiam DL  & I  & 0.67 & 4.3 & 18\% & 0.34 & 5.6 & 23\% & 0.19 & 6.5 & 28\% \\
            \bottomrule
        \end{tabular}
        \begin{tablenotes}
            \item M = meteorological variables (n = 6); I+M = RGB image features (n = 2048) and meteorological variables; I = RGB image features. RMSE units are µg/m\textsuperscript{-3}.
        \end{tablenotes}
    \end{threeparttable}
\end{table}

\begin{landscape}

\begin{table}
    \begin{threeparttable}
        \caption{Bootstrap-estimated 95\% confidence interval of test set performance for the OP\textsubscript{AA}, OP\textsubscript{DTT}, and PM\textsubscript{10} models.}
        \label{sup:tab:boots}
        \small
        \begin{tabular}{llrrrrrrrrr}
            \toprule
            \multicolumn{2}{c}{} & \multicolumn{3}{c}{OP\textsubscript{AA}} & \multicolumn{3}{c}{OP\textsubscript{DTT}} & \multicolumn{3}{c}{PM\textsubscript{10}} \\
            \cmidrule(lr){3-5}
            \cmidrule(lr){6-8}
            \cmidrule(lr){9-11}
            Model & Features & R\textsuperscript{2} & RMSE & NMAE & R\textsuperscript{2} & RMSE & NMAE & R\textsuperscript{2} & RMSE & NMAE \\
            \midrule
            Baseline    & I+M &  0.49 - 0.70 & 0.55 - 0.88 & 0.30 - 0.40 &  0.08 - 0.46 & 0.64 - 0.93 & 0.31 - 0.42 &  0.23 - 0.43 & 5.46 - 6.45 & 0.25 - 0.29 \\
            Random      & I+M &  0.42 - 0.66 & 0.57 - 0.93 & 0.31 - 0.41 &  0.09 - 0.44 & 0.64 - 0.93 & 0.32 - 0.42 &  0.35 - 0.50 & 5.07 - 5.95 & 0.23 - 0.27 \\
            \addlinespace
            Transfer    & I+M &  0.41 - 0.69 & 0.51 - 1.00 & 0.28 - 0.39 &  0.16 - 0.51 & 0.64 - 0.83 & 0.31 - 0.40 &  0.22 - 0.39 & 5.52 - 6.57 & 0.25 - 0.29 \\
            Fine-tuning & I+M &  0.52 - 0.67 & 0.53 - 0.89 & 0.29 - 0.39 &  0.21 - 0.51 & 0.62 - 0.84 & 0.31 - 0.40 &  0.22 - 0.49 & 5.18 - 6.36 & 0.23 - 0.27 \\
            Transfer    & I+M &  0.54 - 0.72 & 0.48 - 0.89 & 0.26 - 0.36 &  0.32 - 0.59 & 0.56 - 0.79 & 0.28 - 0.36 &  0.36 - 0.52 & 4.97 - 5.89 & 0.22 - 0.26 \\
            Fine-tuning & I+M &  0.56 - 0.72 & 0.47 - 0.87 & 0.26 - 0.37 &  0.26 - 0.60 & 0.58 - 0.77 & 0.29 - 0.38 &  0.28 - 0.56 & 4.76 - 6.18 & 0.20 - 0.24 \\
            \addlinespace
            SimSiam     & I+M &  0.27 - 0.61 & 0.60 - 1.08 & 0.34 - 0.47 & -0.05 - 0.40 & 0.70 - 0.95 & 0.35 - 0.46 &  0.20 - 0.36 & 5.67 - 6.66 & 0.26 - 0.30 \\
            SimSiam BJ  & I+M & -0.17 - 0.37 & 0.80 - 1.28 & 0.43 - 0.57 & -0.36 - 0.36 & 0.75 - 1.03 & 0.35 - 0.49 &  0.02 - 0.20 & 6.33 - 7.40 & 0.28 - 0.33 \\
            SimSiam DL  & I+M &  0.20 - 0.55 & 0.68 - 1.04 & 0.37 - 0.50 &  0.02 - 0.47 & 0.67 - 0.88 & 0.32 - 0.43 &  0.08 - 0.23 & 6.15 - 7.27 & 0.27 - 0.32 \\
            \addlinespace
            Random      & I+M &  0.11 - 0.35 & 0.73 - 1.22 & 0.42 - 0.55 &  0.08 - 0.24 & 0.68 - 1.02 & 0.32 - 0.43 &  0.07 - 0.22 & 6.16 - 7.27 & 0.28 - 0.32 \\
            Transfer    & I+M &  0.35 - 0.62 & 0.56 - 1.07 & 0.31 - 0.42 &  0.10 - 0.47 & 0.66 - 0.86 & 0.32 - 0.42 &  0.13 - 0.32 & 5.84 - 6.89 & 0.26 - 0.31 \\
            Fine-tuning & I+M &  0.40 - 0.57 & 0.60 - 1.00 & 0.33 - 0.44 &  0.17 - 0.48 & 0.64 - 0.86 & 0.33 - 0.42 & -0.01 - 0.39 & 5.58 - 7.32 & 0.24 - 0.29 \\
            SimSiam     & I+M & -0.02 - 0.31 & 0.75 - 1.30 & 0.41 - 0.55 & -0.20 - 0.35 & 0.73 - 1.00 & 0.34 - 0.46 &  0.00 - 0.20 & 6.27 - 7.43 & 0.29 - 0.33 \\
            SimSiam BJ  & I+M & -0.07 - 0.37 & 0.75 - 1.29 & 0.42 - 0.57 & -0.37 - 0.19 & 0.79 - 1.10 & 0.39 - 0.52 & -0.03 - 0.18 & 6.51 - 7.44 & 0.29 - 0.35 \\
            SimSiam DL  & I+M &  0.25 - 0.59 & 0.64 - 1.01 & 0.35 - 0.48 & -0.13 - 0.42 & 0.70 - 0.94 & 0.34 - 0.45 &  0.09 - 0.28 & 6.00 - 7.12 & 0.26 - 0.31 \\
                        \bottomrule
        \end{tabular}
        \begin{tablenotes}
            \item M = meteorological variables (n = 6); I+M = RGB image features (n = 2048) and meteorological variables; I = RGB image features. RMSE units are µg/m\textsuperscript{-3}.
        \end{tablenotes}
    \end{threeparttable}
\end{table}

\end{landscape}

\end{document}